\documentclass{article}

\usepackage{microtype}
\usepackage{booktabs}
\usepackage{graphicx}
\usepackage[table]{xcolor}
\usepackage{subcaption}
\usepackage{enumitem}
\usepackage[normalem]{ulem}
\usepackage[most]{tcolorbox}
\usepackage{xspace}
\usepackage{hyperref}

\usepackage[accepted]{mlsys2025}
\makeatletter\renewcommand{\mlsys@appearing}{}\makeatother
\usepackage{cleveref}
\usepackage[labelfont=bf]{caption}

\newcommand{\platform}{GitHub Copilot\xspace}
\newcommand{\myparagraph}[1]{
\vspace{\smallskipamount}
\noindent\textbf{#1.\xspace}}
\newcommand{\myparagraphnodot}[1]{
\vspace{\smallskipamount}
\noindent\textbf{#1\xspace}}
\newcommand{\myparagraphemph}[1]{
\vspace{\smallskipamount}
\noindent\emph{#1.\xspace}}
\newcommand{\eg}{\emph{e.g.}\xspace}
\newcommand{\ie}{\emph{i.e.}\xspace}

\newtcolorbox{simplenoteBox}{colback=green!30!black!7!white, colframe=black, boxrule=0.2mm, arc=1.5mm, auto outer arc, boxsep=0mm, left=2mm, right=2mm, top=1mm, bottom=1mm}
\newcommand{\simplebox}[1]{\begin{simplenoteBox} #1 \end{simplenoteBox}}
\crefname{takeaway}{Takeaway}{Takeaway}
\newcounter{takeaway}
\newcommand{\boxtakeaway}[1]{%
  \refstepcounter{takeaway}%
  \vspace{\smallskipamount}%
  \noindent \simplebox{
    \textbf{\uline{\textit{Takeaway \#~\thetakeaway:}}}
    \textit{#1}%
  }
}

\mlsystitlerunning{Agents in the Wild: Coding Agent Workloads at GitHub Copilot}

\begin{document}

\twocolumn[
\mlsystitle{Agents in the Wild: Characterizing and Optimizing Coding Agents Workload at GitHub Copilot}

\mlsyssetsymbol{intern}{*}

\begin{mlsysauthorlist}
\mlsysauthor{Banruo Liu}{intern,uiuc}
\mlsysauthor{Haoran Qiu}{msr}
\mlsysauthor{Íñigo Goiri}{msr}
\mlsysauthor{Rodrigo Fonseca}{msr}
\mlsysauthor{Ricardo Bianchini}{azure}
\mlsysauthor{Esha Choukse}{msr}
\end{mlsysauthorlist}

\mlsysaffiliation{uiuc}{University of Illinois Urbana-Champaign, Urbana, Illinois, USA}
\mlsysaffiliation{msr}{Microsoft Azure Research, Redmond, Washington, USA}
\mlsysaffiliation{azure}{Microsoft Azure, Redmond, Washington, USA}

\mlsyscorrespondingauthor{Haoran Qiu}{haoran.qiu@microsoft.com}
\mlsyskeywords{AI coding agents, production workload characterization, LLM serving, KV cache, resource orchestration}

\vskip 0.1in

\begin{abstract}
AI coding agents (\eg{}, GitHub Copilot, Claude Code, Codex) interleave multi-step LLM inference with tool execution, creating a workload different from chatbots.
We present the first production-scale characterization of coding agents using sampled \platform traces from June--July 2026, comprising 7.4M users, 819M conversations, 5.7B LLM calls, and 946T tokens.
Our analysis reveals distinctive workload properties with unique systems implications:
\eg{}, sparse user-initiated \textit{turns} unfold into autonomous agent loops with nearly 1:1 coupling between LLM calls and tool execution;
Parallel LLM calls are rare but exists in turns that triggered subagents;
KV cache hit rates average 90\% within a turn, but falls to 52\% across turn boundaries and to 6\% after model switches;
Workflows and user behaviors exhibit diverse, long-tailed resource footprint.
We further distinguish short agentic turnaround times from minutes-long user idle periods between turns, and develop a \textit{lightweight idle-time predictor} that captures 87--89\% of total idle time.
We demonstrate two use cases:
(1) a harness-side LLM API keep-alive policy that preserves provider-side context KV cache, reducing token costs by {14.9\%} on turn boundaries; and
(2) a serving-side KV cache residency policy that uses predicted idle time to guide DRAM offloading, reducing time to first token (TTFT) by {3--33\%} compared with LRU during a one-week evaluation with \platform traces.
\end{abstract}

]

\printAffiliationsAndNotice{\textsuperscript{*}Work done while interning at Microsoft Azure Research.}
\sloppy
\section{Introduction}
\label{sec:intro}

AI coding agents (\eg{}, GitHub Copilot~\cite{copilot}, Claude Code~\cite{claudecode}, and OpenAI Codex~\cite{openaicodex}) are rapidly becoming a dominant class of modern LLM workloads.
These agents autonomously execute a diverse range of software-engineering tasks by exploring codebases, making edits, running builds, diagnosing failures, and iteratively refining solutions.
As a result, a single task may involve dozens of LLM calls interleaved with tool execution over minutes or even \emph{hours}.
Millions of developers now delegate their daily tasks to agents that operate with minimal human intervention (\Cref{tab:dataset}).

Despite this rapid adoption, there is \emph{no empirical understanding} of how coding agents behave at production scale.
LLM serving systems (\eg{}, vLLM~\cite{kwon2023vllm} and SGLang~\cite{zheng2024sglang}) were originally designed for a fundamentally different workload: independent, short-lived, stateless requests.
Recent work has begun to recognize this mismatch~\cite{yu2026pythia,luo2025autellix,lin2024parrot,li2025continuum,biswas2026sutradhara,guo2026saga,chaudhry2025murakkab, liu2026compasssloawarequeryplanner}.
For example, CacheTTL~\cite{li2025continuum} proposes retaining KV-cache state across tool-execution gaps, SAGA~\cite{guo2026saga} introduces workflow-aware scheduling for agent execution graphs, and Sutradhara~\cite{biswas2026sutradhara} co-designs agent orchestrators and inference engines.
However, these systems are designed and evaluated primarily using synthetic benchmarks (\eg{}, SWE-bench~\cite{jimenez2024swebench}), leaving it unclear whether their assumptions and optimizations reflect the behavior of production coding agents.

\begin{table}[t]
\centering
\caption{{Key findings from \platform coding agent traces.} Traces are from {June--July 2026 (sampled 103M agent sessions)}.}
\label{tab:key-findings}
\resizebox{\linewidth}{!}{%
\begin{tabular}{ll}
\toprule
\noalign{\vskip-\belowrulesep}
\rowcolor{gray!15}
\textbf{Finding} & \textbf{Evidence} \\
\noalign{\vskip-\aboverulesep}
\midrule
Coding agents $\neq$ chat & 86\% agent-initiated; 1:1 LLM:tool ratio \\
Cache is session-structured & 90\% cached; $\downarrow$52\% at turn boundaries \\
Model switches destroy cache & only 6\% cached after switch \\
Turn boundaries signal idle & 4.0 (container) and 2.7 (KV cache) minutes \\
Workflows shapes differ & Different LLM/tool call patterns\\
Tool failures amplify cost & 9\% of turns $\to$ 4$\times$ compute via retry loops \\
Users are highly heterogeneous & 50\texttimes{} token range across 5 user archetypes \\
\bottomrule
\end{tabular}
}%
\vspace{-10pt}
\end{table}

\myparagraph{Characterizing Production Workloads}
We fill this gap with the \emph{first production-scale characterization} of AI coding agents.
Our dataset comprises sampled traces from \platform's production coding agent, spanning 103.7M sessions and 819.3M conversations from 7.4M users during the two-month period of June and July 2026 and encompassing 5.7B LLM calls, 946.3T total prompt and completion tokens, and 7.2B tool invocations.
This characterization uncovers many interesting observations and structural properties of production coding-agent workloads that challenge conventional LLM serving assumptions.
For example:
\begin{enumerate}[leftmargin=*,nosep,topsep=0pt]
\vspace{-0.5\baselineskip}
\item \textbf{Self-initiated LLM$\leftrightarrow$Tool coupling.}
Agent execution forms an autonomous loop of interleaved LLM calls and tool invocations, with {86\%} of LLM calls initiated by the agent.
This tight coupling calls for session-aware scheduling beyond independent request optimization.

\item \textbf{Session-structured KV-cache lifecycle.}
Prefix caching covers an average of {90\%} of tokens within turns, but cache hit rates drop to {52\%} across turn boundaries (due to idleness) and just {6\%} after model switches.
These sharp variations make KV cache retention, eviction, and offloading key scheduling decisions.

\item \textbf{Heterogeneous workflows and users.}
Five distinct user archetypes span a 50$\times$ range in token consumption per-turn (23K to 1.1M tokens), alongside distinct tool usage, idle-time profiles, and session statefulness. These differences make uniform resource allocation inefficient.

\item \textbf{Tool failures amplify compute costs.}
Tool failures affect {9\%} of turns and can trigger autonomous retry loops that amplify compute by up to {4$\times$}. Tool reliability and retry behavior are therefore important cost factors.

\item \textbf{Actionable idle periods.}
Agent execution alternates between GPU-bound inference and CPU/I/O-bound tool execution, creating short idle gaps within turns and minute-long pauses at turn boundaries.
These predictable idle periods offer opportunities for proactive container reclamation and KV-cache residency management.
\end{enumerate}
\noindent
\Cref{tab:key-findings} summarizes these findings.
While some of these observations might not be surprising, our production traces quantify their prevalence and system-level impact at scale, revealing patterns that synthetic benchmarks alone cannot establish.
Together, these findings provide an empirical foundation for evaluating serving system designs and developing optimizations grounded in real-world agent behavior.

\myparagraph{Idle-Time Prediction and Use Cases}
Building on these observations, we develop a lightweight idle-time predictor (\Cref{sec:idle-predictor}) that captures {87--89\%} of total idle time.
We demonstrate its effectiveness in two independent use cases:
(1) a harness-side LLM API keep-alive policy (\Cref{sec:keepalive}) that reduces token costs by {14.9\%} at turn boundaries, and
(2) a serving-side KV-cache residency policy (\Cref{sec:offloading}) that reduces TTFT latency by {3--33\%} compared with LRU cache eviction policy during a one-week evaluation.

\myparagraph{Contributions}
In summary, our main contributions are:
\begin{itemize}[leftmargin=*,nosep,topsep=0pt]
\vspace{-0.4\baselineskip}
\item The largest production-scale characterization\footnote{Open-sourced to support reproducibility and future agentic systems research. \url{https://github.com/Azure/AzurePublicDataset/blob/master/GitHubCopilotCodingAgentDataset2026.md}} of LLM coding-agent workloads, quantifying LLM-call and tool-execution patterns that distinguish agentic workloads from conventional LLM serving (\Cref{sec:agent-exec,sec:llm,sec:tools}).
\item A quantitative analysis of KV-cache lifecycle dynamics across agent trajectories, establishing cache residency as a session-aware scheduling problem (\Cref{sec:kv-cache}).
\item An empirical characterization of workflow and user heterogeneity, revealing substantial variation in resource demands and execution patterns (\Cref{subsec:workflows,sec:users}).
\item A lightweight idle-time predictor and two applications to resource management, reducing costs by {14.9\%} and TTFT by {3--33\%} on real-world traces (\Cref{sec:idle-predictor,sec:keepalive,sec:offloading}).
\end{itemize}

\section{Coding Agents in Production}

\subsection{Background on Coding Agents}
\label{sec:background}

AI coding agents~\cite{claudecode,openaicodex,copilot} extend conversational LLMs with the ability to take actions in a software developer environment.
Building on the ReAct paradigm~\cite{yao2023react}, modern foundation models are trained~\cite{feng2025retool} to invoke tools and reason over tool outputs, enabling them to perform tasks such as code migration, testing, and terminal interaction.

In frontier AI platforms~\cite{copilot,claudecode,openaicodex}, this agentic execution loop runs with minimal human intervention.
Tools may execute directly on the user's workspace or within isolated sandbox environments~\cite{claudecodesandbox,jain2025livecodebench} depending on the operation and security requirements.
A single user request can therefore trigger dozens of LLM calls and tool invocations, forming a workflow that spans both GPU-backed inference and CPU/IO-intensive tool execution.

\begin{table}[t!]
\caption{{Chatbot vs.\ Agentic Coding workloads.}
}
\label{tab:chat-vs-agent}
\centering
\resizebox{\linewidth}{!}{%
\begin{tabular}{lll}
\toprule
\noalign{\vskip-\belowrulesep}
\rowcolor{gray!15}
\textbf{Property} & \textbf{Chat/Completion} & \textbf{Coding Agent} \\
\noalign{\vskip-\aboverulesep}
\midrule
Token asymmetry & Moderate & Extreme ({341:1 In:Out}) \\
State between calls & Stateless/replay & Tight sequential dependency \\
Resource pattern & GPU only & GPU $\leftrightarrow$ CPU/IO alternation \\
Session duration & Seconds & Seconds to minutes or hours \\
Failure handling & N/A (manual retry) & Retry loops ({48$\times$ P95 blowup}) \\
Cache sensitivity & Low-to-medium & High (98\% median cache hit) \\
Autonomy level & User-driven & Agent-driven (86\% LLM calls) \\
\bottomrule
\end{tabular}
}%
\end{table}

\myparagraph{Distinction from Chat}
The agentic coding workload differs from single-turn chat (and even multi-turn conversation)~\cite{gao2024cost} in several structural ways that impact serving infrastructure.
These differences, summarized in \Cref{tab:chat-vs-agent}, mean that serving infrastructure designed for chat (\eg{}, request-level scheduling and LRU cache eviction) is structurally mismatched to coding-agent workloads.

\subsection{Dataset and Methodology}
\label{sec:methodology}

\begin{table}[t!]
    \caption{{Statistics of the dataset scale and coverage.}}
    \label{tab:dataset}
    \centering
    \resizebox{\linewidth}{!}{%
    \begin{tabular}{cccc}
    \toprule
    \noalign{\vskip-\belowrulesep}
    \rowcolor{gray!15}
    \textbf{Sessions} &
    \textbf{Turns} &
    \textbf{\# Users} &
    \textbf{\# Models / Tools} \\
    \noalign{\vskip-\belowrulesep}
    \midrule
    103.7M & 819.3M & 7.4M & 92+ / 197+ \\
    \midrule
    \noalign{\vskip-\belowrulesep}
    \rowcolor{gray!15}
    \textbf{LLM Calls} &
    \textbf{Tool Calls} &
    \textbf{Prompt Tokens} &
    \textbf{Output Tokens} \\
    \noalign{\vskip-\belowrulesep}
    \midrule
    5.7B & 7.2B & 942.2T & 4.1T \\
    \bottomrule
    \end{tabular}
    }%
\end{table}

\myparagraph{At-Scale Production Data Source}
We derive our dataset from anonymized telemetry collected from \platform's coding agent during {June and July 2026}.
The telemetry captures system metadata for every LLM call and tool invocation, including timestamps, durations, input/output token counts, model names, tool names, and success or failure status.
Overall, the traces include 103.7M sessions and 819.3M turns from 7.4M individual users, covering at least 92 unique models and 197 tools. They encompass 5.7B LLM calls and 7.2B tool calls, with 942.2T prompt tokens and 4.1T completion tokens (\Cref{tab:dataset}).
To protect user confidentiality, we analyze a sampled subset of these anonymized traces.
We do not collect the prompt text, model outputs, source code, tool arguments, or any user-identifiable information.

\myparagraph{Session Hierarchy}
Traces follow a three-level hierarchy:
\begin{itemize}[leftmargin=*,nosep,noitemsep]
\item \textbf{Session}: a coding agent session lifetime.
\item \textbf{User-Turn} (or simply \textbf{Turn}): one agentic coding \emph{conversation} triggered by the user prompt and followed by the agent's full autonomous response chain.
\item \textbf{Step}:  a single LLM invocation or tool call within a turn; the last step in a turn is always an LLM invocation.
\end{itemize}

Within each turn, the canonical execution pattern is:
\begin{center}
\footnotesize
\texttt{User prompt $\to$ LLM $\to$ [LLM | tools $\to$ LLM]$^*$ $\to$ final response back to user}
\end{center}
Every turn begins with exactly one user-initiated LLM call; all subsequent LLM invocations or tool calls within the turn are agent-initiated (auto-continued).

\begin{figure}[t]
\centering
\begin{subfigure}[b]{0.49\linewidth}
    \centering
    \includegraphics[width=\linewidth]{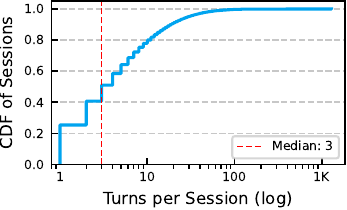}
    \caption{Number of turns per session.}
    \label{fig:session-turns-cdf}
\end{subfigure}
\begin{subfigure}[b]{0.49\linewidth}
    \centering
    \includegraphics[width=\linewidth]{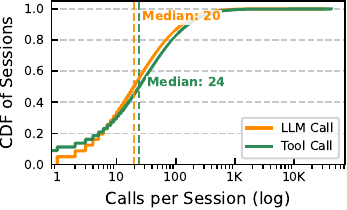}
    \caption{Number of steps per session.}
    \label{fig:session-calls-cdf}
\end{subfigure}
\begin{subfigure}[b]{0.49\linewidth}
    \centering
    \includegraphics[width=\linewidth]{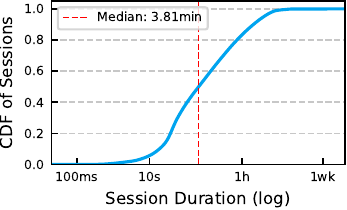}
    \caption{Session duration.}
    \label{fig:session-duration-cdf}
\end{subfigure}
\hfill
\begin{subfigure}[b]{0.49\linewidth}
    \centering
    \includegraphics[width=\linewidth]{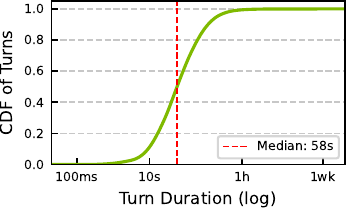}
    \caption{Turn duration.}
    \label{fig:turn-duration-cdf}
\end{subfigure}
\begin{subfigure}[b]{0.99\linewidth}
    \centering
    \includegraphics[width=\linewidth]{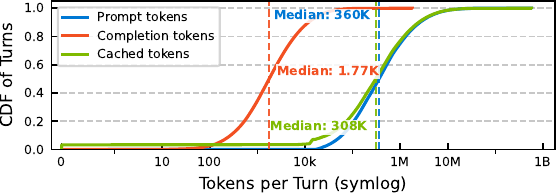}
    \caption{Token consumption per turn.}
    \label{fig:turn-tokens-cdf}
\end{subfigure}
\caption{Distributions of session and turn metrics.
}
\label{fig:duration-distributions}
\end{figure}

\begin{table}[t]
\caption{{Summary of session/turn statistics to \Cref{fig:duration-distributions}.} ``Skew'' measures the mean-to-median ratio of each metric.}
\label{tab:session-overview}
\centering
\resizebox{\linewidth}{!}{%
\begin{tabular}{lrrrrr}
 \toprule
 \noalign{\vskip-\belowrulesep}
 \rowcolor{gray!15}
 \textbf{Metric} & \textbf{Mean} & \textbf{Median} & \textbf{P75} & \textbf{P90} & \textbf{Skew} \\
 \noalign{\vskip-\aboverulesep}
 \midrule
 User turns per session    & 7.9   & 3      & 8      & 19      & 2.6$\times$ \\
 LLM calls per turn        & 7.0   & 3      & 7      & 16      & 2.3$\times$ \\
 Tool calls per turn       & 8.8   & 3      & 10     & 23      & 2.9$\times$ \\
 Input tokens per turn     & 1.15M & 359.8K & 1.00M  & 2.62M   & 3.2$\times$ \\
 Cached tokens per turn    & 1.07M & 307.8K & 905.6K & 2.45M   & 3.5$\times$ \\
 Output tokens per turn    & 5.0K  & 1.8K   & 4.8K   & 11.9K   & 2.8$\times$ \\
 Turn duration (s)         & 303.5 & 58.0   & 163.5  & 415.2   & 5.2$\times$ \\
 \bottomrule
 \end{tabular}
}%
\end{table}

\subsection{Session/Turn-Level Overview}
\Cref{fig:duration-distributions} shows the distributions of key session- and turn-level metrics, including LLM calls, tool invocations, token consumption, and duration;
\Cref{tab:session-overview} summarizes these statistics across the collection period.
Session characteristics exhibit strong heavy-tailed behavior.
A median session contains only 3 user turns, 20 LLM calls, and lasts 3.8 minutes, whereas the mean reaches 7.9 turns, 55.4 LLM calls, and 52.5 minutes, highlighting substantial right skew.
At the 90$^{th}$ percentile, they reach 19 user turns, nearly 150 LLM and tool invocations.
Session duration is especially skewed, with a 13.8$\times$ mean-to-median ratio and a P95 $>$4 hours, although most sessions complete within 1 hour (\Cref{fig:session-duration-cdf}).

At the turn level within each session, the median turn triggers 3 LLM calls and 3 tool invocations, consuming 359.8K prompt tokens and 1.8K completion tokens.
However, the upper tail is substantial:
P90 turns require nearly 16 LLM calls and 23 tool invocations, and over 2.6M prompt tokens.
\Cref{fig:turn-duration-cdf} shows that the turn durations are likewise highly skewed (5.2$\times$): most turns finish within 10 minutes while a minority of complex turns dominate execution time and token consumption, even for solving a single task.

\subsection{Temporal Patterns}
\label{sec:diurnal}

Our dataset covers the two-month period of June and July 2026. The time-series figures below focus on a one-week subset to illustrate temporal patterns at production scale.

\begin{figure}[t!]
\centering
\includegraphics[width=\linewidth]{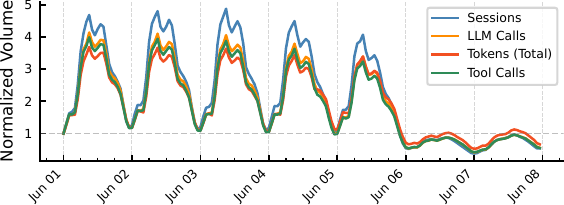}
\caption{Traffic trend from a subset of traces over {one week in June 2026} within the two-month collection period.
All metrics are sampled from the full population and normalized to day~1 hour~1.
}
\label{fig:volume-trends}
\end{figure}

\begin{figure}[t!]
\centering
\includegraphics[width=\linewidth]{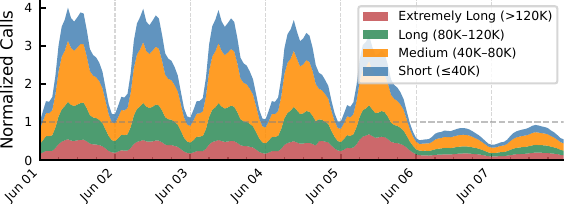}
\caption{LLM call volume by prompt length category.
}
\label{fig:prompt-trends}
\end{figure}

\myparagraph{Aggregated Trends}
\Cref{fig:volume-trends} shows the normalized workload volume during the observation period.
All aggregate metrics, including sessions, LLM calls, tool calls, and total-token volume, exhibit highly synchronized behavior and strong diurnal patterns.
Activity peaks during daytime hours, reaching approximately $4$--$5\times$ the baseline level, and declines substantially overnight (and during the weekend).
\Cref{fig:prompt-trends} further decomposes call volume by prompt length category. While all bins follow the same diurnal rhythm, they do not vary equally: extremely long and long prompts remain comparatively stable across time, whereas medium-length prompts show the largest swings in volume, driving most of the overall diurnal fluctuation.

\begin{figure}[t!]
\centering
\includegraphics[width=0.95\linewidth]{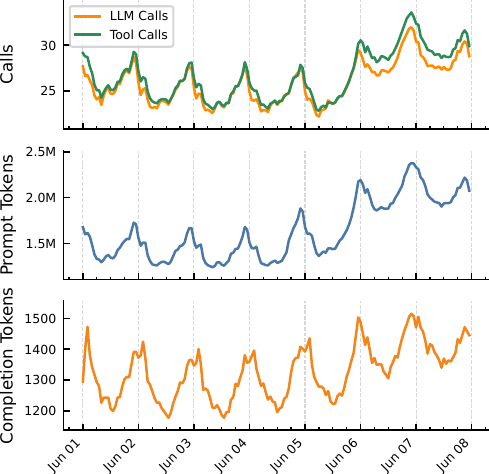}
\caption{Per-session average metrics from {sampled traces} during the same one-week window in June 2026.
}
\label{fig:session-avg-trends}
\end{figure}

\myparagraph{Per-Session Average Trends}
To examine whether coding-agent session patterns change over time, \Cref{fig:session-avg-trends} reports per-session average statistics.
During weekdays, sessions contain approximately 22--28 LLM calls and 23--29 tool calls on average, while weekend sessions are noticeably more intensive, increasing to roughly 27--33 LLM calls and 28--34 tool calls per session.
Prompt-token usage rises from about 1.2M--1.7M tokens during weekdays to 1.9M--2.4M on weekends, while average completion-token usage increases more modestly by 0.1K.
We observe the same diurnal and weekday-versus-weekend patterns across P25/P50/P75 percentiles, indicating that this shift toward fewer, more intensive sessions reflects a broad change in typical session behavior.

\begin{figure}[t]
\centering
\includegraphics[width=\linewidth]{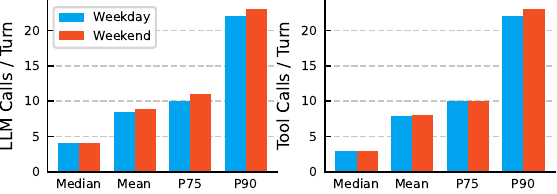}
\caption{Comparison of the number of LLM invocations and tool calls per turn during weekdays vs. weekends.
}
\label{fig:weekday-weekend-cdf}
\end{figure}

This heavy-tail is further modulated by temporal patterns.
Weekend sessions are fewer but longer (\Cref{fig:session-avg-trends}), with more LLM/tool calls per turn, suggesting developers might undertake more ambitious tasks when uninterrupted (\Cref{fig:weekday-weekend-cdf}).
This differs from prior observations in chatbots, where longer sessions are more common during weekdays~\cite{anthropic-study}.

Next, we present a deeper analysis of the sessions, LLM calls, tool calls, agentic workflow patterns, and users in the subsequent sections.

\section{Agent Execution Pattern}
\label{sec:agent-exec}

We first characterize the fundamental execution pattern and workflows observed in coding agent workloads.

\subsection{Agent Execution Structure}
\label{sec:exec-structure}

\myparagraph{Session and Turn Composition}
About 3/4 of sessions contain multiple turns, but these sessions account for {96.8\%} of all turns (\Cref{tab:session-composition}).
Multi-turn sessions are also more likely to combine LLM calls with tool invocations ({94.3\%}) than single-turn sessions ({81.3\%}), indicating that longer sessions tend to involve more interactive agent execution.

\begin{table}[t!]
    \centering
    \small
    \caption{Session structure and execution composition.}
    \label{tab:session-composition}
    \begin{tabular}{lrr}
        \toprule
        \noalign{\vskip-\belowrulesep}
        \rowcolor{gray!15}
        \textbf{Metric} & \textbf{Single-turn} & \textbf{Multi-turn} \\
        \noalign{\vskip-\aboverulesep}
        \midrule
        Share of sessions & 25.4\% & 74.6\% \\
        Share of turns & 3.2\% & 96.8\% \\
        \midrule
        Both LLM and tool calls & 81.3\% & 94.3\% \\
        LLM calls only & 18.7\% & 5.7\% \\
        \bottomrule
    \end{tabular}
\end{table}

\begin{figure}[t]
    \centering
    \includegraphics[width=\linewidth]{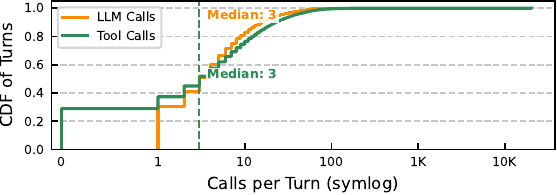}
    \caption{{CDFs of per-turn LLM calls and tool invocations.}
    }
    \label{fig:per-turn-cdf}
\end{figure}

\myparagraph{The \texorpdfstring{{1:1} LLM$\leftrightarrow$Tool}{1:1 LLM-Tool} Ratio}
Across the full population, the ratio of LLM calls to tool invocations is remarkably close to {1:1} (median {20} vs.\ {24} per session, \Cref{fig:session-calls-cdf}).
This observation also holds at the \emph{per-turn} level, as shown in \Cref{fig:per-turn-cdf}.
The median turn triggers nearly identical numbers of LLM and tool calls, and the CDFs of per-turn LLM calls and tool invocations closely track each other across the entire distribution, except for those LLM-only turns.

This tight coupling reflects the canonical \texttt{observe} $\rightarrow$ \texttt{(reason) decide} $\rightarrow$ \texttt{act} loop of agentic systems~\cite{yao2023react}.
Most LLM calls result in a tool action whose results immediately trigger another LLM call autonomously.
As a result, the agent autonomously executes an average of {6} LLM calls after a user message, yielding a split of approximately {86\%} agent-initiated vs. {14\%} user-initiated calls.
The distribution is highly skewed (\Cref{fig:per-turn-cdf}): a small fraction of requests trigger long execution chains that account for a disproportionate share of LLM serving load.

\boxtakeaway{Agentic serving exhibits strong {1:1} LLM$\leftrightarrow$tool coupling, with {86\%} of LLM calls initiated by autonomous agents. Serving systems must model interdependent LLM-tool interactions at the turn level, rather than treating requests independently or relying solely on user request arrivals for capacity planning.}

\begin{figure}[t!]
  \centering
  \begin{subfigure}[t]{0.48\linewidth}
    \centering
    \includegraphics[width=\linewidth]{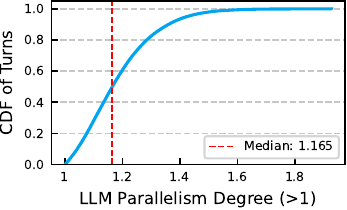}
    \caption{LLM call parallelism.}
    \label{fig:llm-concurrency}
  \end{subfigure}
  \hfill
  \begin{subfigure}[t]{0.48\linewidth}
    \centering
    \includegraphics[width=\linewidth]{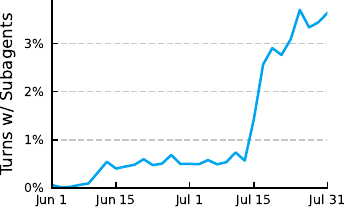}
    \caption{Turns with subagents.}
    \label{fig:turn-subagent-pct}
  \end{subfigure}
  \caption{Measured LLM call parallelism and subagent usage.}
  \label{fig:parallelism-and-subagents}
\end{figure}

\myparagraph{Parallel LLM Calls and Subagents}
Although agentic workflows can spawn parallel subtasks, LLM execution is overwhelmingly serial.
\Cref{fig:llm-concurrency} shows the distribution of per-turn LLM concurrency.
Across all turns, {38.6\%} are strictly sequential, while the remaining {61.4\%} exhibit limited overlap between LLM calls (measured by the overlap of their wall clock times).
However, the degree of parallelism is shallow: the median concurrency is only {1.17}, and the {P90} reaches {1.38}.
The limited concurrency stems from the dependency structure of agent execution. Most LLM calls consume outputs produced by earlier steps.
When concurrency does occur, it typically arises from independent subtasks or subagents (\eg{}, background agents) that can run simultaneously.
The parallelism might increase as we observe an uptrend of triggering subagents (\Cref{fig:turn-subagent-pct}).

\myparagraph{Temporal Patterns of LLM Call Concurrency}
\begin{figure}[t]
    \centering
    \includegraphics[width=\linewidth]{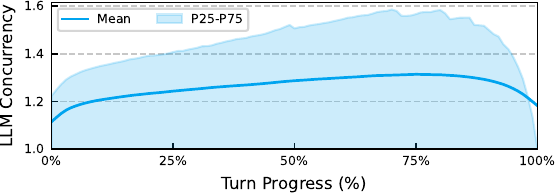}
    \caption{{LLM parallelism over turn progress.}
    }
    \label{fig:per-turn-llm-parallelism}
\end{figure}

LLM concurrency follows an inverted-U pattern over turn progress (\Cref{fig:per-turn-llm-parallelism}).
Turns typically begin with a single reasoning step, after which agents may issue multiple concurrent requests during exploration or information gathering.
As execution converges toward a final decision, concurrency declines toward one because downstream reasoning depends on the results of preceding branches.
Thus, concurrency is concentrated in the middle of turns, while their beginnings and endings remain largely sequential.

This pattern creates two serving challenges. First, stragglers can delay downstream LLM calls or tool invocations when an agent must wait for all parallel upstream calls to finish. Second, serving systems must support concurrent KV-cache access and updates for multiple calls from the same session.

\begin{figure}[t]
    \centering
    \includegraphics[width=\linewidth]{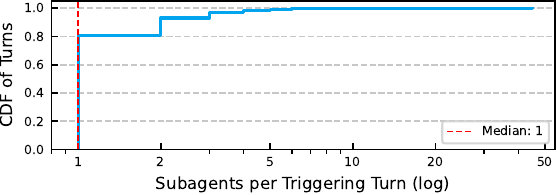}
    \caption{{CDFs of number of subagents per turn.}
    }
    \label{fig:per-turn-subagents}
\end{figure}

The degree of parallelism is also related to subagent spawning (\Cref{fig:per-turn-subagents}): most turns trigger only one subagent, while a small fraction spawn several (with a long tail triggering 20+), providing opportunities for parallel execution.

\begin{figure}[t]
    \centering
    \includegraphics[width=\linewidth]{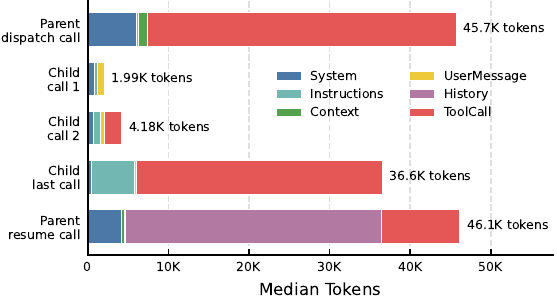}
    \caption{{Context composition of parent and subagent LLM calls for a median scenario.}
    }
    \label{fig:context-subagents}
\end{figure}

\myparagraph{Subagent Context Composition}
\Cref{fig:context-subagents} compares the median context composition of parent and subagent LLM calls. The parent dispatch call contains 45.7K tokens, dominated by tool-call content, while child calls use substantially smaller contexts (1.99K--36.6K tokens).
This reflects a common pattern: subagents receive only the context relevant to their assigned tasks, rather than the full parent context.
When resuming the parent agent execution, the parent receives a summary of the subagent's output rather than its full execution history, which contributes to the 46.1K-token context dominated by accumulated conversation history.

\boxtakeaway{Agentic execution is predominantly serial. While {61\%} of turns exhibit some overlap, concurrency remains shallow ({P90=1.4}) and concentrated in the middle of turns, creating potential straggler dependencies and same-session KV-cache contention.}

\subsection{Agentic Workflow Archetypes}
\label{subsec:workflows}
Each agentic turn is a dynamically generated workflow based on its specific task content.
To better understand workflow diversity, we cluster turns using their tool composition, LLM-call depth, token consumption, and other execution characteristics.
This analysis reveals six major, recurring workflow archetypes shown in \Cref{tab:archetypes}.

\begin{table}[t]
\caption{Median turn-level workflow archetypes.}
\label{tab:archetypes}
\centering
\resizebox{0.97\linewidth}{!}{%
\begin{tabular}{lrlr}
\toprule
\noalign{\vskip-\belowrulesep}
\rowcolor{gray!15}
\textbf{Archetype}   & \textbf{LLM calls}  & \textbf{Description} & \textbf{Share} \\
\noalign{\vskip-\aboverulesep}
\midrule
Deep-loop read       & 9  & 7 tool batches; read-heavy & 30.5\% \\
LLM-only             & 1  & No tools; pure reasoning & 20.2\% \\
Multi-cycle edit     & 5  & Read + edit + build & 19.0\% \\
Multi-cycle other    & 4  & Read-dominant & 13.2\% \\
Deep-loop w/failures & 36 & 34 tool batches; retry loops & 9.1\% \\
Deep-loop run        & 7  & Terminal-heavy & 8.1\% \\
\bottomrule
\end{tabular}
}%
\end{table}

\myparagraphemph{Exploration and Reasoning}
The largest category, \textbf{Deep-loop read} ({30.5\%}), consists of extended code exploration involving repeated file retrieval, symbol lookup, and repository navigation. These turns average {9} LLM calls and {7} tool batches as the agent gathers context before making modifications. In contrast, \emph{LLM-only} turns ({20.2\%}) contain a single LLM call without tool usage, corresponding to pure reasoning, planning, or explanatory interactions.

\myparagraphemph{Modification and Execution}
\textbf{Multi-cycle edit} turns ({19.0\%}) alternate between reading, modifying, and validating code, often forming short feedback loops between edits and build or test execution. \textbf{Deep-loop run} turns ({8.1\%}) are dominated by terminal interactions, with repeated command execution and reasoning over the resulting output. The \textbf{Multi-cycle other} category ({13.2\%}) involves multiple rounds of exploration and reasoning without substantial code changes.

\begin{figure}[t!]
\centering
\includegraphics[width=.8\linewidth]{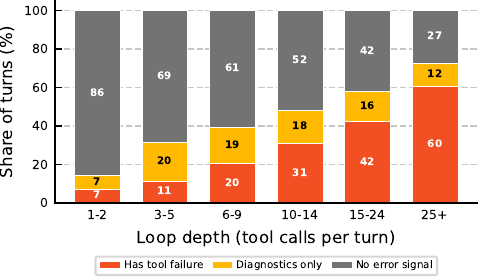}
\caption{Tool failure frequency in turns of varying lengths.}
\label{fig:failure-turn-size}
\end{figure}

\myparagraphemph{Failure-driven Retries}
The remaining {9.1\%} of turns, classified as \textbf{Deep-loop w/failures}, contain extended debugging and retry sequences. These turns have a median of {36} LLM calls--four times the median of that from turns without failures--and process {80K+} tokens per call, including accumulated error output, across {34} tool batches. This failure-driven \emph{compute amplification} is distinctive to agentic workloads: instead of simply returning an error, a failed operation can trigger autonomous diagnosis and recovery, cascading into additional LLM calls and growing context windows.
\Cref{fig:failure-turn-size} illustrates a positive correlation between tool failures and turn lengths.

\begin{figure}[t!]
\centering
\includegraphics[width=0.78\linewidth]{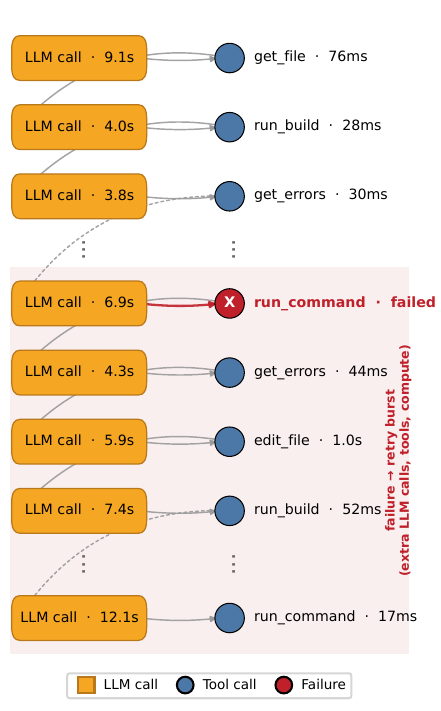}
\caption{Timeline of a ``Deep-loop w/failures'' turn: repeated tool failures trigger autonomous retry loops, generating {36} LLM calls and {35} tool calls in total in this example.}
\label{fig:failure-timeline}
\end{figure}

As \Cref{fig:failure-timeline} illustrates, agents may encounter tool failures like failed builds, incorrect assumptions, missing dependencies, or unexpected execution results. They then alternate between reasoning, tool execution, diagnosis, and corrective actions before converging on a successful solution.
This trial-and-error behavior distinguishes agentic workloads from conventional request-response interactions and can substantially increase resource consumption.

This diversity creates substantial variability in serving demand: turns differ widely in LLM/tool calls, and token consumption depending on the amount of exploration, validation, and failure recovery required.
Efficient serving therefore requires \emph{workflow-aware} scheduling that accounts for the evolving resource demands of entire workflows, rather than treating LLM calls as independent requests.

\boxtakeaway{Coding-agent workflows are highly heterogeneous, resulting in a large variation in call-dependency and token consumption. Iterative retry workflows can amplify cost by {4}$\times$, making workflow-aware scheduling important for efficient serving.}

\section{LLM Footprints and KV-Cache}
\label{sec:llm}

We next characterize LLM inference footprints (\Cref{sec:llm-basics}) and their implications for KV-cache efficiency (\Cref{sec:kv-cache}), focusing on two session-structural events: \textbf{turn boundaries} (\Cref{sec:kv-cache}) and \textbf{model switches} (\Cref{sec:model-switches}).

\subsection{LLM Call Basics}
\label{sec:llm-basics}

\begin{figure}[t!]
\centering
\includegraphics[width=\linewidth]{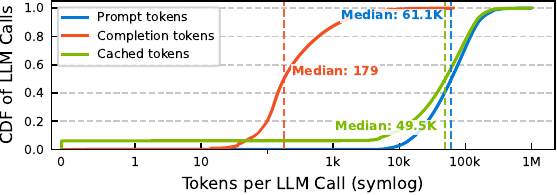}
\caption{CDFs of per-LLM-call token counts.}
\label{fig:token-cdfs}
\end{figure}

\myparagraph{Token Length Distribution}
\Cref{fig:token-cdfs} shows the CDFs of per-call token consumption.
Coding-agent workloads are strongly input-heavy:
The median prompt contains {61K} tokens, including {49.5K} cached prompt tokens, whereas the median completion contains only {179} tokens, and {88\%} of calls produce fewer than {1K} output tokens.
The ratio of median prompt to median completion length exceeds {341:1}, shifting the serving bottleneck from token generation to KV-cache efficiency, a challenge we quantify in \Cref{sec:kv-cache}.

\begin{figure}[t!]
\centering
\includegraphics[width=\linewidth]{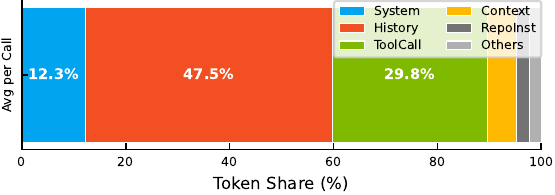}
\caption{Token type breakdown for LLM calls.}
\label{fig:token-type-breakdown}
\end{figure}

\Cref{fig:token-type-breakdown} breaks down input tokens by source. Conversation history contributes {47.5\%} of input tokens on average, followed by function-call messages ({29.8\%}) and system prompts ({12.3\%}).
Repository instructions and other contextual information account for the remaining $\sim${10\%}.

\myparagraph{Difference Compared to Chat}
Compared with production traces from text-only and multimodal LLM APIs~\cite{stojkovic2025dynamollm,modserve2024qiu}, coding-agent calls have substantially longer inputs and outputs.
Their median prompt length of {61K} tokens far exceeds the {750} and {1,050} tokens reported for text-only and multimodal workloads, respectively.
Median completions in coding agents are also longer: {179} tokens versus {105} and {80}, respectively. This difference reflects the extensive code, execution logs, and accumulated context processed during agent execution.

\boxtakeaway{Coding-agent calls are highly token-intensive and substantially longer than those observed in text and multimodal LLM API traces.}

\begin{figure}[t!]
\centering
\includegraphics[width=\linewidth]{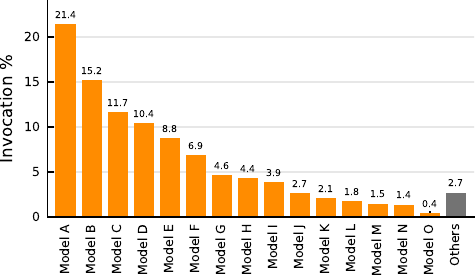}
\caption{Distribution of inference usage of top 15 models.
}
\label{fig:model-distribution}
\end{figure}

\myparagraph{Model Coverage}
Our traces span a diverse set of at least 92 underlying models that stem from a variety of model families, reflecting the heterogeneity of models used by coding agents in production.
\Cref{fig:model-distribution} shows the distribution of inference usage across the top 15 models observed during the study period.
The distribution is heavily skewed: the top three models (Model A, B, and C) together account for approximately {48.3\%} of all invocations, with Model A alone responsible for {21.4\%}.
Usage then falls off sharply, with the remaining twelve models each contributing {0.4--10.4\%} of invocations, and the long tail of additional models collectively accounting for the remaining {2.7\%}.

\subsection{LLMs Dominating Execution Time}
\label{sec:time-breakdown}

\begin{figure}[t!]
\centering
\begin{subfigure}[b]{\linewidth}
    \centering
    \includegraphics[width=\linewidth]{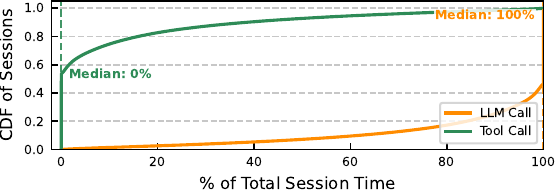}
    \caption{Single-turn sessions.}
    \label{fig:time-breakdown-single-turn}
\end{subfigure}
\hfill
\begin{subfigure}[b]{\linewidth}
    \centering
    \includegraphics[width=\linewidth]{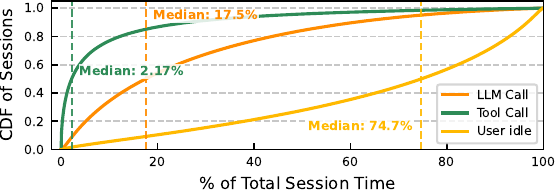}
    \caption{Multi-turn sessions.}
    \label{fig:time-breakdown-multi-turn}
\end{subfigure}
\caption{CDF of session wall-clock time spent in LLM execution, tool execution, and (across-turns) user idle periods.
}
\label{fig:time-breakdown-cdf}
\end{figure}

\Cref{fig:time-breakdown-cdf} breaks down session wall-clock time into three categories: (1) LLM execution, (2) tool execution, and (3) user idle time between turns.
Among all sessions, {25.4\%} are single-turn sessions with no user-idle time between turns.
For those sessions with both LLM and tool calls, LLM inference dominates runtime with a median of {87.7\%}.
For multi-turn sessions, user-idle time becomes the major contribution to the total wall clock time, with a median of {74.7\%}.
Among the rest of the session runtime, LLM inference time still dominates tool call with a median of {17.5\%} vs. {2.2\%}.

Overall, agentic coding sessions are fundamentally \emph{LLM-bound} workloads.
Optimizing model serving latency directly improves end-to-end completion time for the vast majority of sessions, whereas tool-system optimizations primarily benefit the relatively small fraction of workflows dominated by long-running external operations.

\boxtakeaway{Agentic sessions are overwhelmingly LLM-bound, but the time and token contributions are inverted. LLM call takes {88\%} of wall-clock time yet in medium contributes 179 tokens per call, whereas tool calls take only 1/7 of time yet contribute 3\texttimes{} more tokens.}

\subsection{KV Cache Lifecyles}
\label{sec:kv-cache}

Within a turn, successive LLM calls reuse an increasingly long context, enabling effective prefix caching. However, two session-structural events---\textbf{turn boundaries} and \textbf{model switches}---can substantially reduce cache reuse.

\begin{figure}[t]
\centering
\begin{subfigure}[b]{0.47\linewidth}
    \centering
    \includegraphics[width=\linewidth]{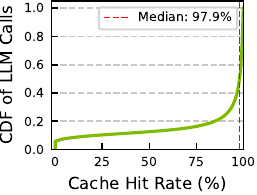}
    \caption{CDF of cache hit rate.}
    \label{fig:cache-hit-cdf}
\end{subfigure}
\hfill
\begin{subfigure}[b]{0.47\linewidth}
    \centering
    \includegraphics[width=\linewidth]{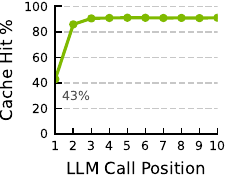}
    \caption{Cache hit rate progression.}
    \label{fig:cache-by-step}
\end{subfigure}
\caption{Cache hit rate analysis.}
\label{fig:cache-hit-distribution}
\end{figure}

\myparagraph{Prefix Cache Hit Rate}
As conversation history accumulates within a turn, successive calls extend an increasingly reusable prompt prefix.
Accordingly, the median cache hit rate reaches {97.9\%}.
However, \Cref{fig:cache-hit-cdf} reveals a bimodal distribution: roughly {10\%} of LLM calls have hit rates below {25\%}, while over {80\%} exceed {90\%}.
The low-hit-rate calls largely reflect cold starts, whereas the majority achieve near-complete prefix reuse.

The cache hit rate follows a predictable trajectory over the sequence of LLM calls within a turn (\Cref{fig:cache-by-step}):
\begin{enumerate}[leftmargin=*,nosep]
\item \textbf{Call {\#1}}: $\sim${43\%} (cold start, first call has the minimal prefix reuse).
\item \textbf{Call {\#2}}: rises to $\sim${86\%} as turn history becomes reusable.
\item \textbf{Call {\#3} onward}: stablizes at $\sim${93--98\%} (new content added per call offsets the prefix reuse gains).
\end{enumerate}

\boxtakeaway{Prefix caching rate is high overall, and follows a predictable trajectory within a turn, rising from {43\%} on initial call to {93--98\%} on subsequent calls.}

\begin{figure}[t]
\centering
\includegraphics[width=\linewidth]{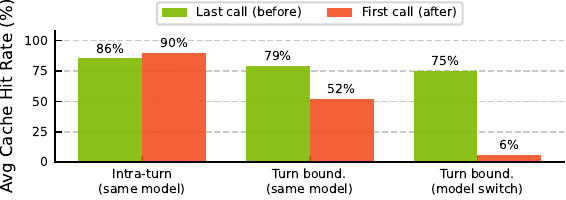}
\caption{Cache hit rate in LLM calls before and after turn boundaries. Same-model boundaries cause a $\downarrow${27\%} average drop; model-switch boundaries cause near-complete invalidation ($\downarrow${69\%}).}
\label{fig:turn-boundary-cache}
\end{figure}

\myparagraph{Cache Degradation at Turn Boundaries}
A turn boundary separates one autonomous response chain from the next user message.
As shown in \Cref{fig:turn-boundary-cache}, these boundaries are a major source of cache degradation: the average hit rate drops by {27\%} when consecutive turns use the same model, and by {69\%} when the model changes.
The gap between turns often includes user idle time, during which cached entries may be evicted by the serving system.

\begin{figure}[t!]
\centering
\includegraphics[width=\linewidth]{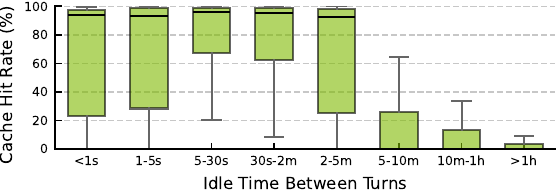}
\caption{Cache hit rate drops with inter-turn idle time.}
\label{fig:turn-idle-cache}
\end{figure}

\Cref{fig:turn-idle-cache} quantifies the relationship between inter-turn idle time and cache reuse.
For gaps $<${2} minutes, median hit rates remain at or above {95\%}.
At the 5-min boundary, median falls to $\sim${70\%}, and the distribution widens considerably, with some sessions losing nearly all cached state.
Beyond {10} minutes, median hit rates drop to $<${5\%}, and the upper quartile rarely exceeds {20\%}.
This sharp decline after several minutes is consistent with time-based KV-cache eviction.

\boxtakeaway{Turn boundaries degrade absolute cache hit rates by $\downarrow$27\% on average, primarily via time-based serving-system eviction during inter-turn idle periods.}

\subsection{Model Switches}
\label{sec:model-switches}
\begin{figure}[t!]
\centering
\includegraphics[width=\linewidth]{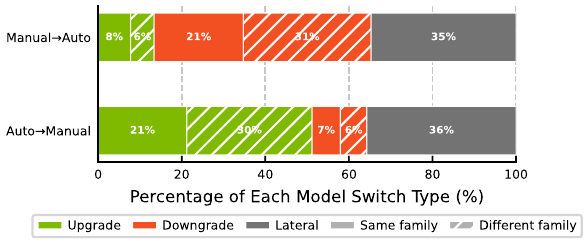}
\caption{Composition of model switches by direction.
The bars show the share that upgrade to a higher-tier model, downgrade to a lower-tier model, or move laterally to a same-tier model based on model sizes and generations.}
\label{fig:model-switches-results}
\end{figure}

Since a session's model is fixed for the duration of a turn, model switches only occur at turn boundaries, and when they do, they destroy the cache entirely rather than merely degrading it: a KV cache built for one model's weights and attention layout cannot be reused by another model.
We observe that model switches affect $\sim${6.4\%} of sessions and are predominantly \emph{reactive}, triggered by errors ({36\%} non-success rate before switches vs.\ {8\%} baseline) or by rate limiting.
Switches arise from two sources: (1) explicit user intent (the user selects a different model for the next turn) and (2) auto mode, where the system itself reroutes the session, primarily in response to rate limiting or throttling on the currently selected model.
\Cref{fig:model-switches-results} breaks down switches by direction (size and family) and outcome.

Manual-to-auto switches are dominated by downgrades ({52\%}), with lateral moves accounting for {35\%} and upgrades only {13\%}, consistent with auto mode stepping a session down to a cheaper or more available model once the user relinquishes explicit control.
Auto-to-manual switches show the opposite pattern: {51\%} are upgrades, as users who take back manual control tend to move to a stronger model, while {36\%} are lateral and only {13\%} are downgrades.

After a model switch, the average cache hit rate is only {6\%}, a $-${69\%} average drop compared to $-${27\%} for same-model turn boundaries (as shown in \Cref{fig:turn-boundary-cache}); this gap isolates the pure cost of model incompatibility, on top of the eviction loss that all turn boundaries incur.

Since each model switch effectively cold-starts the cache, pinning sessions to a single model would preserve cache continuity.
When switches are unavoidable (\eg{}, throttling-driven), optimization is needed to precompute and stage the new model's KV cache before switching rather than incurring a synchronous cold-start which imposes significant latency and cost spikes on top of the KV cache eviction loss already incurred at the turn boundary.

\boxtakeaway{Model switches are mostly reactive to rate limiting, compound turn boundary into near-total cache loss ($\downarrow${69\%}, average hit rate {6\%}). Session-to-model pinning and proactive cache staging on the target model are needed to avoid this added cold-start cost.}

\section{Tool Execution}
\label{sec:tools}

Although tool calls consume relatively little wall-clock time overall, their outputs contribute substantially to prompt tokens (\Cref{sec:time-breakdown}), while their latency and failure behavior determine how long agents wait within a turn.

\begin{figure}[t!]
\centering
\includegraphics[width=.92\linewidth]{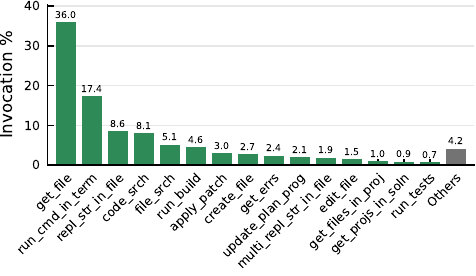}
\vspace{-8pt}
\caption{Tool invocation frequency.}
\vspace{-10pt}
\label{fig:tool-usage}
\end{figure}

\myparagraph{Tool Profile}
At least {197} distinct tools are observed being used from the traces, but their usage is heavily concentrated (\Cref{fig:tool-usage}).
A single tool, \texttt{get\_file}, accounts for {36.0\%} of all invocations, followed by \texttt{run\_cmd} ({17.4\%}) and \texttt{replace\_string} ({8.6\%}).
The top {11} tools alone account for over {90\%} of invocations, and the top {25} for {99\%}, with the long remaining tail of tools together contributing just {4.2\%} (mostly customized tools).

\begin{figure}[t!]
\centering
\includegraphics[width=\linewidth]{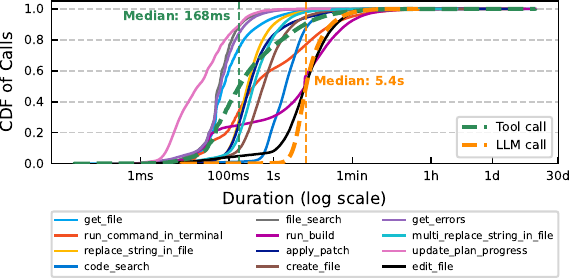}
\caption{Distribution of tool call vs. LLM call duration, overall and for the 12 most-invoked tools.}
\label{fig:tool-llm-times-combo}
\end{figure}

\myparagraph{Execution Latency and Diversity}
Tool execution times are highly right-skewed and substantially shorter than LLM calls (\Cref{fig:tool-llm-times-combo}): the median is only {168\,ms} versus {5.4\,s} for LLM calls, but the mean reaches {13.9\,s}, with P90 and P99 latencies of {5\,s} and {98.5\,s}, respectively.
The mean is nearly {83$\times$} the median as a small fraction of long-running calls dominate total tool execution time.
\Cref{fig:tool-llm-times-combo} further reveals a {320$\times$} spread in median latency across the {12} most-invoked tools.
Read operations are consistently fast and dominate usage, whereas mutation tools have intermediate latency and execution-oriented tools drastically stretch the tail.
Thus, infrequent execution calls can introduce substantial delays despite accounting for a small share of invocations.

\begin{figure}[t!]
\centering
\vspace{-10pt}
\includegraphics[width=\linewidth]{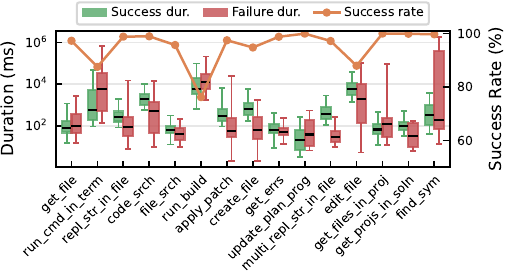}
\vspace{-20pt}
\caption{Tool execution duration by success/failure status.}
\label{fig:tool-duration-status}
\end{figure}

\myparagraph{Failure Behavior}
Tool reliability also varies substantially by type (\Cref{fig:tool-duration-status}).
Read and search operations, such as \texttt{get\_file} and \texttt{code\_search}, succeed nearly {100\%} of the time, whereas execution and mutation tools, including \texttt{run\_cmd}, \texttt{run\_build}, and \texttt{edit\_file}, have success rates of only {70--85\%}.
Failed calls also exhibit longer and more variable execution times; for \texttt{run\_cmd}, the P95 latency of failed calls is {48$\times$} that of successful calls.
These failures extend blocking intervals before the agent can process tool results and resume LLM inference.

\begin{figure}[t!]
\centering
\includegraphics[width=\linewidth]{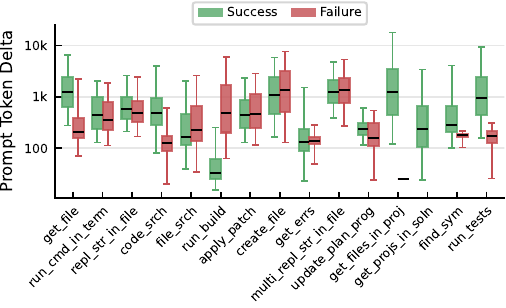}
\caption{Tokens added from tool call outputs grouped by success/failure status of tool execution.}
\label{fig:tool-tokens-status}
\end{figure}

\myparagraph{Added Prompt Tokens from Tool Outputs}
Beyond wall-clock time, tool invocations also differ substantially in how many prompt tokens they inject into the agent's context, and this cost is not uniform across success and failure states (\Cref{fig:tool-tokens-status}).
For most read-oriented and mutation tools such as \texttt{get\_file}, \texttt{code\_search}, and \texttt{apply\_patch}, successful and failed calls contribute a broadly similar number of tokens, with failures occasionally trending lower since error responses are often terse compared to full file or search contents.
One clearest exception to this pattern is \texttt{run\_build}: successful builds return minimal output (a median of roughly {60} tokens), whereas failed builds inject substantially more context, often {7--8$\times$} more at the median, consistent with compilers emitting verbose diagnostics, stack traces, or error logs on failure.
This compounds the latency effect described above---failed \texttt{run\_build} calls are both slower and more context-expensive, doubly taxing the agent loop.
The opposite pattern appears for \texttt{run\_tests}, where successful runs report richer token deltas (\eg{}, coverage or per-test summaries) while failures return less output.

\begin{figure}[t]
    \centering
    \includegraphics[width=\linewidth]{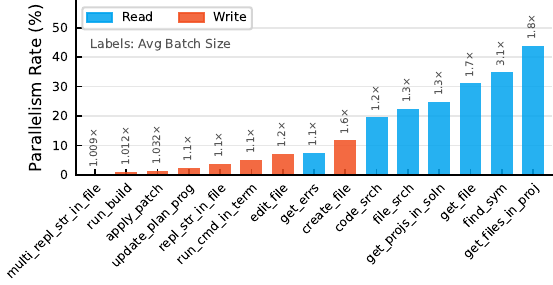}
    \vspace{-20pt}
    \caption{Per-tool parallelism categorized into R/W actions.}
    \label{fig:per-tool-parallelism}
    \vspace{-5pt}
\end{figure}

\myparagraph{Tool Call Parallelism}
Tool calls are predominantly sequential, with parallelism concentrated in read-only operations.
\Cref{fig:per-tool-parallelism} shows that tools such as \texttt{get\_file} and \texttt{get\_symbols} are often invoked together, enabling concurrent information gathering (\eg{}, code understanding).
In contrast, mutation and execution tools, such as \texttt{run\_cmd} and \texttt{run\_build}, are rarely parallelized, as shared-state changes and side effects often require serialization.

\begin{figure}[t!]
    \centering
    \includegraphics[width=\linewidth]{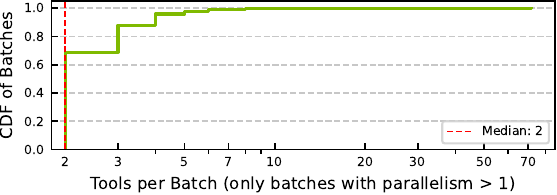}
    \caption{Tool call concurrency during a turn.
    {92\%} of batches contain a single tool (issued in batches), but a power-law tail extends to {70+} parallel tools.
    }
    \label{fig:tool-parallelism}
\end{figure}

\Cref{fig:tool-parallelism} shows the distribution of per-turn tool parallelism.
Across all tool batches, {92\%} contain a single tool invocation, while only {8\%} dispatch multiple tools concurrently.
Among the parallel batches, the median width is {2} tools and {87.5\%} contain at most {3} tools, indicating that most parallelism is shallow despite a long tail that reaches {70+} concurrent tool invocations.

The prevalence of small read-only batches suggests that agents already exploit parallelism when exploring code bases, but only conservatively.
Given that information-gathering phases account for a substantial fraction of turns, more batching of independent read operations could reduce wall-clock latency without introducing consistency risks.
At the same time, the rarity of large parallel batches indicates that existing serving systems need not optimize for extreme tool fan-out in the common case.

\boxtakeaway{Tool execution is more parallel than LLM execution, dominated by a small set of frequently used tools, but infrequent, long-running, and failure-prone execution calls drive tail latency.}

\begin{figure}[t!]
\centering
\includegraphics[width=\linewidth]{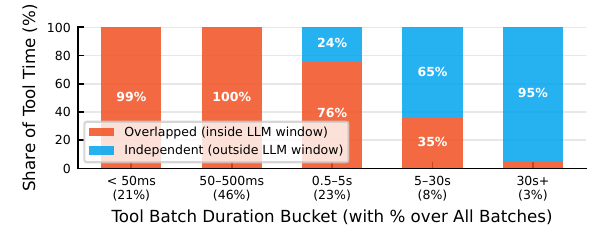}
\caption{Fraction of tool execution time that overlaps with an active LLM call, grouped by tool-batch duration.}
\label{fig:tool-llm-parallelism}
\end{figure}

\myparagraph{LLM-Tool Overlap}
LLM-tool parallelism~\cite{biswas2026sutradhara,xu2024conveyor} has been applied to hide tool execution fully or partially behind ongoing LLM inference.
\Cref{fig:tool-llm-parallelism} partitions tool-batch execution time into \emph{overlapped} intervals (executing while at least one LLM call is active) and \emph{independent} intervals (executing outside any LLM call time window).

Short tool batches are almost completely shadowed by LLM execution.
For batches shorter than {500\,ms}, over {99\%} of tool time overlaps with an active LLM call.
Even for {0.5--5\,s} batches, {76\%} of execution time remains hidden.
Only long-running tool batches become visible on the critical path: for {5--30\,s} batches, {65\%} of tool time occurs independently of LLM execution, while for {30\,s+} batches nearly all ({95\%}) tool time lies outside LLM windows. We believe the overlap comes from techniques like async tool calling.

This overlap pattern reveals a sharp gap between per-call and aggregate behavior.
By batch \emph{count}, overlap is pervasive: {97\%} of tool batches run at least partially inside an LLM window, and the many short-lived information-gathering calls are almost fully hidden behind ongoing model inference.

In terms of the aggregate \emph{wall-clock time}, however, overlap barely dents tool execution: only {7.7\%} of total tool time is actually hidden inside LLM windows, while the remaining {92\%} sits on the critical path.
This is because the long-tail of large, long-running tools, such as builds, test executions, or external commands, dominates total tool wall-clock time and is largely \emph{not} overlapped, materially affecting the session wall-clock time that end users perceive.

\boxtakeaway{Tool-LLM overlap is pervasive by count ({97\%} of batches run inside an LLM window) but hides only {7.7\%} of aggregate tool wall-clock time; the long-tail of long-running tools dominates total tool time, is largely un-overlapped, and drives session latency.}

\section{User Heterogeneity}
\label{sec:users}

\begin{table*}[t!]
\caption{User types depending on the coding agent usage. Values are per-user medians.}
\label{tab:user-types}
\centering
\resizebox{0.85\linewidth}{!}{%
\begin{tabular}{lrrrrrl}
\toprule
\noalign{\vskip-\belowrulesep}
\rowcolor{gray!15}
\textbf{User type} & \textbf{\% Users} & \textbf{Sessions} & \textbf{Turns} & \textbf{Tools per Turn} & \textbf{Tokens per Turn} &\textbf{Description} \\
\noalign{\vskip-\aboverulesep}
\midrule
Readers    & 41.7 & 2 & 6 & 4.8 & 203K & Mostly reads/searches \\
Coders   & 30.4 & 5 & 50 & 6.2 & 417K & Balanced read+edit+exec \\
Terminal users   & 11.0 & 2 & 7  & 4.0 & 213K & Mostly terminal calls \\
Deep-loop users  &  9.2 & 2 & 6  & 20.0 & 1.1M & Long agentic loops \\
Chat-only users  &  7.6 & 1 & 2  & 0.0 & 23K & 100\% Q\&A turns; no tools \\
\bottomrule
\end{tabular}
}%
\end{table*}

Linking sessions through anonymized user identifiers lets us characterize how the usage is distributed across the developer population, how different user behaviors translate into distinct resource footprint usage over time, and what these differences imply for serving-system SLOs.

\subsection{User Archetypes}
\Cref{tab:user-types} groups users into five behavioral archetypes based on their per-turn tool/llm-call mix and token consumptions (values are per-user medians).

\myparagraph{Readers ({41.7\%})}
The largest group runs short sessions (median {2} sessions, {6} turns) dominated by file reads and code searches, consuming a modest {203K} tokens per turn (median).
These users primarily use the coding agent for information retrieval, \eg{}, exploring unfamiliar codebases, looking up API signatures, or gathering context before making a decision.
Their sessions are fast and stateless from a resource perspective: fast and parallel tool execution (read tools complete in tens of milliseconds), slightly longer tool outputs, and overall short session durations.

\myparagraph{Coders ({30.4\%})}
The most engaged group by session volume, with a median of {5} sessions and {50} turns per user during the observation period.
Coders balance reads, edits, and executions at {417K} tokens per turn and {6.2} tools per turn, reflecting the full software-engineering workflow: gather context, modify code, validate via build or test.

\myparagraph{Terminal Users ({11.0\%})}
These users lean heavily on command execution (\texttt{run\_cmd\_in\_terminal}), with a median of {4.0} tools per turn and {213K} tokens per turn.
Their sessions are characterized by highly variable tool/command latency: terminal commands range from near-instant to minutes-long builds or test suites, creating unpredictable idle patterns that complicate resource scheduling.

\begin{figure}[t!]
\centering
\includegraphics[width=\linewidth]{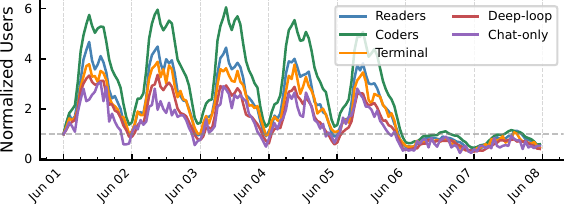}
\caption{Users of different types varying with time.
}
\label{fig:user-types-distribution}
\end{figure}

\myparagraph{Deep-loop Users ({9.2\%})}
The most resource-intensive user type, running long autonomous loops with {20.0} tools per turn and {1.1M} tokens per turn---5$\times$ the token consumption of readers and 2.6$\times$ that of coders.
Despite having only {2} median sessions and {6} turns per user, each turn generates substantial serving load due to the extended chains of LLM$\leftrightarrow$tool interactions.
These users delegate complex, multi-step tasks (large refactors, cross-file migrations, debugging sessions) and rely on the agent's autonomy to iterate without intervention.
As shown in Figure~\ref{fig:user-types-distribution}, the deep-loop users show more activity during the weekends compared to weekdays, as compared to the other archetypes. This leads to overall higher tool calls per turn during the weekends.

\myparagraph{Chat-only Users ({7.6\%})}
These users never invoke tools, issuing pure question-and-answer turns at just {23K} tokens per turn.
With a median of {1} session and {2} turns, they represent the lightest workload---closer to a traditional chatbot interaction than an agentic coding workflow.

\begin{figure}[t!]
\centering
\includegraphics[width=\linewidth]{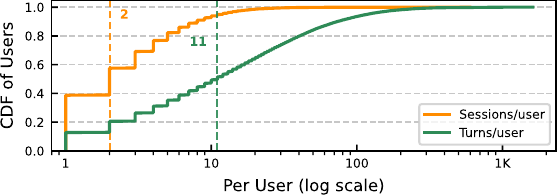}
\caption{Distribution of the number of sessions and turns per user during June 1 to June 7.
}
\vspace{-15pt}
\label{fig:user-turns-distribution}
\end{figure}

\subsection{Resource Concentration}
\label{sec:resource-concentration}

\begin{figure}[t!]
\centering
\includegraphics[width=\linewidth]{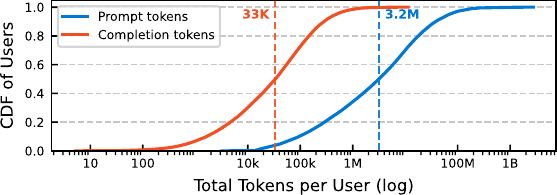}
\caption{Distribution of total prompt token and completion token consumption per user.
}
\vspace{-20pt}
\label{fig:user-tokens-distribution}
\end{figure}

Consumption is heavily concentrated across users.
\Cref{fig:user-turns-distribution} shows that the per-user session and turn counts are strongly right-skewed: most users run only a handful of sessions (median of 2 sessions per user, P90 of 8), while a small fraction of highly active developers accumulate hundreds of turns (median of 11 turns per user, P90 of 74).
\Cref{fig:user-tokens-distribution} shows the same concentration in token consumption, where total prompt tokens dominate completion tokens by orders of magnitude (median of 3.2M vs.\ 33K tokens per user, respectively) and a minority of users account for a disproportionate share of the total token volume (P90 of 38M prompt tokens and 282K completion tokens per user).
Together, these distributions indicate that a small subset of intensive users (\ie{}, Coders and Deep-loop users) drives the bulk of serving load, echoing the heavy-tailed behaviors.

\subsection{Implications for Serving Infrastructure}
The cost of a cache miss (eviction followed by re-prefill) varies by over an order of magnitude across archetypes: for Deep-loop users, the per-turn median has {1.1M} tokens, whereas for Chat-only users each turn only consumes {23K} tokens.
This 50$\times$ disparity in per-miss cost means that a uniform KV-cache eviction timeout imposes disproportionate a latency tax on the most resource-intensive user segments.
Beyond cache cost, Coders and Terminal users accumulate significant session state (modified files, running processes, environment variables, build artifacts) within their tool-execution containers, making premature container reclamation expensive; Readers and Chat-only users maintain minimal state and can be cold-started with negligible overhead.
Together, these differences motivate a \emph{tiered SLO model} for agent serving:
\begin{enumerate}[leftmargin=*,nosep,topsep=0pt]
    \item \textbf{Retention priority}: Deep-loop and Coder sessions should receive higher KV-cache retention priority due to their large re-prefill cost and higher probability of near-term reuse compared to the other types.
    \item \textbf{Eviction aggressiveness and container lifecycle}: Chat-only and Reader sessions can be evicted after short idle timeouts, without meaningful latency penalty, freeing memory for higher-priority sessions. Terminal and Coder users require stateful container management (checkpointing rather than termination), while Reader containers can be cold-started freely.
    \item \textbf{Capacity planning}: The heavy-tailed user distribution means that per-user fair-share policies must account for the 50$\times$ token-consumption gap between Chat-only and Deep-loop users to avoid both starvation of intensive users and over-provisioning for light users.
\end{enumerate}

\boxtakeaway{
User archetypes span a 50$\times$ range in per-turn token consumption, making uniform resource policies suboptimal. Archetype-aware SLOs (\eg{}, longer cache retention for Deep-loop users, aggressive eviction for Chat-only/Reader sessions) can reduce tail latency for power users while freeing aggregate memory.}

\section{Resource Idleness Prediction}
\label{sec:idle-predictor}

\subsection{Idle Time Distributions}
\label{subsec:idle-time}

Agentic workflows alternate among LLM inference, tool execution, and user input, leaving different resources temporarily unused.
We systematically quantify these idleness opportunities for resource reclamation.

\begin{itemize}[leftmargin=*,nosep]
\item \textbf{Container idle}: the elapsed time between two consecutive tool invocations. During this interval, the agent is processing tool outputs through one or more LLM calls, leaving the tool-execution container idle.
\item \textbf{KV-cache idle}: the elapsed time between two consecutive LLM calls. During this interval, the agent is executing tools or waiting for user input, while the corresponding KV cache remains allocated but unused.
\end{itemize}

\begin{figure}[t]
\centering
\begin{subfigure}[b]{0.49\linewidth}
    \centering
    \includegraphics[width=\linewidth]{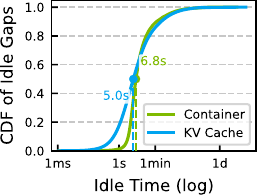}
    \caption{CDF of KV cache and (tool) container idle time.}
    \label{fig:idle-kv-container-cdf}
\end{subfigure}
\hfill
\begin{subfigure}[b]{0.49\linewidth}
    \centering
    \includegraphics[width=\linewidth]{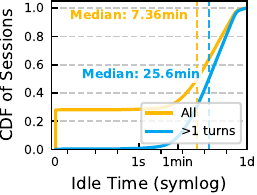}
    \caption{CDF of user idle time (that happens between two turns).}
    \label{fig:idle-user-cdf}
\end{subfigure}
\caption{Idle time distribution.}
\label{fig:idle-cdf}
\vspace{-10pt}
\end{figure}

\Cref{fig:idle-kv-container-cdf} shows the CDF of both idle intervals observed across all sessions.
Container idle times have a median of {6.8\,s}, reflecting the time required for the LLM to process tool outputs and decide the next action.
In contrast, KV-cache idle times are shorter at the median ({5.0\,s}), and nearly one-third of consecutive LLM calls occur with negligible idle time between them due to lightweight tool invocations, allowing KV cache reuse with little delay.

\myparagraph{User Idle Time between Turns}
Both distributions in \Cref{fig:idle-kv-container-cdf} exhibit heavy tails extending to hours, mainly caused by gaps that span \textit{turn boundaries}, where the agent waits for user input before continuing the workflow.
\Cref{fig:idle-user-cdf} shows the distribution of such user idle times.
The median user idle time is 7.36 minutes, with a long tail extending beyond one day (users picking up an unattended session the next day), indicating that user activity time dominates the largest idle intervals observed.

\myparagraph{Intra-turn vs. Cross-Turn Idleness}
To separate autonomous agent execution from user-driven pauses, we classify each idle interval as either \emph{intra-turn} (entirely within an agent turn) or \emph{cross-turn} (spanning a turn boundary).
Table~\ref{tab:idle} shows the resulting breakdown.
More than {90\%} of idle intervals are intra-turn, with median durations of only {5.8\,s} for containers and {1.2\,s} for KV caches.
In contrast, the relatively small fraction of cross-turn intervals (8--10\%) have median durations of 241\,s and 160\,s, respectively, over two orders of magnitude larger.
Unlike the cumulative user-idle time per session shown in \Cref{fig:idle-user-cdf}, \Cref{tab:idle} reports individual resource-idle intervals, so the corresponding percentiles are not directly comparable.

Therefore, turn boundaries provide a natural signal for resource lifecycle management.
Intra-turn idle periods are typically too short to justify sleeping containers or evicting KV caches, as reclamation overheads would frequently exceed the idle duration itself.
Cross-turn idle periods, however, are long enough to amortize state migration and restoration costs, making them attractive opportunities for container hibernation and KV-cache offloading.

\begin{table}[t]
\caption{Resource idle time distributions.}
\label{tab:idle}
\centering
\small
\resizebox{0.95\linewidth}{!}{%
\begin{tabular}{@{}lrrrc@{}}
\toprule
& \multicolumn{2}{c}{\textbf{Intra-Turn}} & \multicolumn{2}{c}{\textbf{Cross-Turn}} \\
\cmidrule(lr){2-3} \cmidrule(lr){4-5}
\textbf{Resource} & \textbf{P50} & \textbf{P95} & \textbf{P50} & \textbf{P95} \\
\midrule
Container & 5.8s & 46s & 241s (4.0 min) & 5{,}575s ($\sim$93 min) \\
KV cache & 1.2s & 38s & 160s (2.7 min) & 4{,}431s ($\sim$74 min) \\
\bottomrule
\end{tabular}
}%
\end{table}

\boxtakeaway{Intra-turn idle periods are short whereas cross-turn idle periods are minutes long due to user idle time. Turn boundaries therefore provide a natural trigger for container eviction and KV-cache offloading.}

\subsection{Turn Boundaries as Reclamation Signals}
\label{subsec:predictor}

We study whether idle periods in agentic workflows are predictable enough to guide resource reclamation, focusing on \textit{KV cache} due to their high cost.
We distinguish cross-turn \emph{user idleness} from \emph{tool-driven idless} and find that user idleness offers substantially greater savings potential.

\myparagraph{User Idleness}
Serving systems can decide whether to retain or reclaim the session's KV cache at turn boundaries.
The decision depends on whether the upcoming idle period is long enough to amortize state migration and reload costs.

\begin{figure}[t!]
\centering
\begin{subfigure}[b]{0.97\linewidth}
    \centering
    \includegraphics[width=\linewidth]{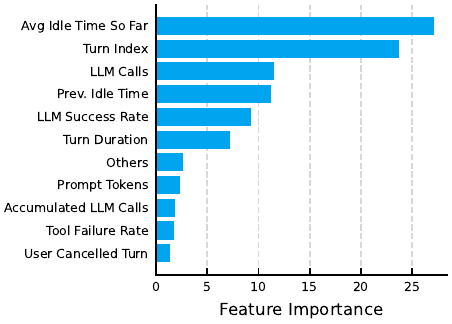}
    \caption{Feature importance for user-idleness prediction.}
    \label{fig:idle-predictor-feature}
\end{subfigure}
\hfill
\begin{subfigure}[b]{\linewidth}
    \centering
    \includegraphics[width=\linewidth]{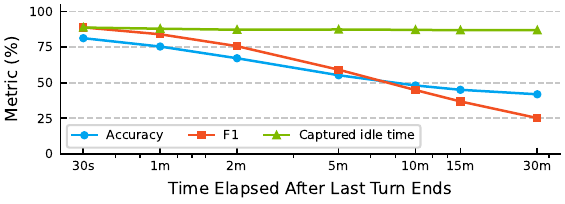}
    \caption{Fixed-horizon classification metrics.}
    \label{fig:idle-predictor-accuracy}
\end{subfigure}
\caption{User-idleness prediction trained on June 1--7 and evaluated on June 8--14. The classification threshold maximizes mean F1 on June 1--7. Captured time measures idle duration beyond each target horizon.}
\label{fig:idle-predictor}
\end{figure}

\myparagraphemph{Input Features}
Each completed turn is summarized by \emph{turn-level} features (duration, LLM calls, tokens, tool time, failures, success rate), \emph{session-level} features (turn index, elapsed time, prior idle gap, average idle so far), and \emph{temporal} features (\eg{}, \texttt{is\_weekday}, \texttt{hour\_in\_day}).
Session-level features dominate: average idle duration and turn index account for over half of feature importance, followed by the previous idle duration and LLM success rate (\Cref{fig:idle-predictor-feature}).

\myparagraphemph{Prediction Model}
Let $T$ denote the idle duration and $S(t)=\Pr(T>t)$ its survival function.
We estimate $S(t)$ using LightGBM quantile regressors over the idle time, trained on completed-turn metadata.
After $t_0$ seconds of idleness, the probability of remaining idle for another $\Delta$ seconds is:
\begin{equation}
\vspace{-10pt}
\Pr(T>t_0+\Delta\mid T>t_0)
={S(t_0+\Delta)}/{S(t_0)}.
\end{equation}

\myparagraphemph{Training Details}
We train on 882K turns from June 1--7 and evaluate on 740K turns from June 8--14.
The quantile ensemble contains 12 models, occupies about 14\,MB, and evaluates in under 1.5\,ms per turn boundary.
LightGBM is within 0.009 ROC-AUC of eight alternative architectures, and richer workflow features improve ROC-AUC by at most 0.003 (\Cref{app:user-idle-prediction}).

\myparagraphemph{Prediction Accuracy}
We evaluate fixed-horizon classifiers that predict whether the idle gap $T$ exceeds a horizon $h$ at turn completion, and measure the fraction of eligible idle time captured beyond $h$:
\begin{equation}
\mathrm{Captured}(h)=
\frac{\sum_{i:T_i>h,\ \hat y_i(h)=1}(T_i-h)}
     {\sum_{i:T_i>h}(T_i-h)}.
\end{equation}
An always-positive classifier achieves 100\% on this metric, so we interpret it alongside precision, which reflects the frequency of false-positive reclamation decisions.
Across horizons, the classifiers recall 86.0--86.5\% of long gaps and capture 86.9--88.8\% of eligible idle time (\Cref{fig:idle-predictor-accuracy}), indicating that they identify most of the potentially reclaimable time.
The proportion of gaps exceeding the horizon falls from 87\% at 30 seconds to 11\% at 30 minutes, explaining the corresponding precision decline from 0.92 to 0.15.

\myparagraphemph{Feature Ceiling and Session-End Prediction}
User idle duration depends on developer activities outside the agent, including meetings, breaks, and task switching, which are absent from telemetry.
No turn feature has a correlation above 0.28 with session termination, limiting the prediction of long-tail idleness.
We therefore train a separate \emph{last-turn} classifier on the same features that predicts whether a completed turn is the session's last (\Cref{app:last-turn-prediction-setup}).
It reaches a ROC-AUC of 0.76 on held-out sessions, and its outputs are calibrated probabilities.

We demonstrate two use cases of the trained idleness predictor in (1) extending LLM API keep-alive windows (\Cref{sec:keepalive}) and (2) KV cache offloading (\Cref{sec:offloading}).

\myparagraph{Tool-driven Idleness}
During tool execution, the session's KV cache remains idle. Unlike user idleness, tool duration depends largely on the agent's actions and is therefore easier to predict.
We predict each tool call's duration at dispatch with a LightGBM regressor over tool identity, batch shape, turn-start context, and completed prior batches (\Cref{app:tool-idle-prediction}).
It improves the within-2$\times$ rate from 61.8\% for a per-tool median baseline to 68.9\% (\Cref{tab:tool-prediction}).
Prediction errors are concentrated in the long tail: terminal commands have a median duration of 0.51\,s but a P99 of 514\,s, and calls lasting at least 60\,s account for 75.6\% of the absolute prediction error.
Results on two external traces show the same tail-dominated error (\Cref{app:tool-idle-prediction}).

\begin{table}[t!]
\centering
\caption{Tool-duration prediction on \platform traces.}
\label{tab:tool-prediction}
\resizebox{\linewidth}{!}{%
\begin{tabular}{@{}lrrrrr@{}}
\toprule
\textbf{Model} & \textbf{MAE (s)} & \textbf{Median AE (s)} & \textbf{RMSE (s)} & $R^2$ & \textbf{Within 2$\times$} \\
\midrule
Per-tool median & 6.041 & 0.117 & 86.334 & -0.002 & 61.8\% \\
LightGBM & 5.712 & 0.092 & 84.786 & 0.034 & 68.9\% \\
\bottomrule
\end{tabular}}
\end{table}

However, tool idleness offers very limited reclamation opportunities.
Of held-out tool calls, 77\% finish within one second; file reads have a median duration of 68\,ms and string edits 260\,ms.
Only 1.31\% of calls last at least 60\,s and 0.32\% at least five minutes, compared with 75\% of user-idle gaps exceeding one minute.
Although the longest 1\% of calls account for 68.6\% of total tool time, tools occupy only a small fraction of session time: in multi-turn sessions, their median share is 2\%, compared with 74.7\% for user idleness (\Cref{sec:time-breakdown}).
In addition, short tool calls largely overlap with LLM execution (\Cref{sec:tools}).

\begin{figure}[t!]
\centering
\includegraphics[width=\linewidth]{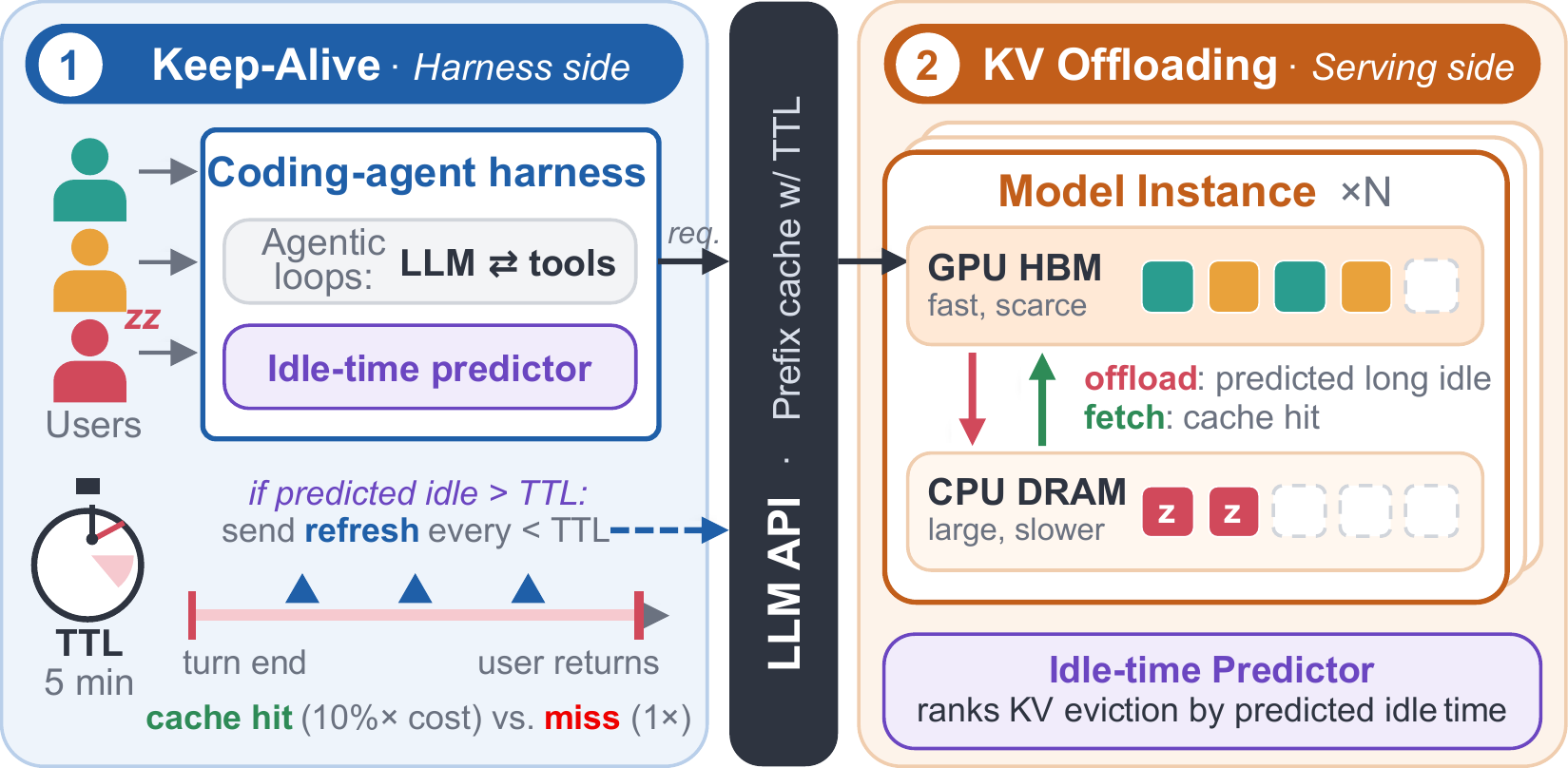}
\caption{Two independent use cases: Idleness-informed session context keep-alive (left) and KV-cache offloading (right).}
\label{fig:use-cases}
\end{figure}

\section{Idleness-Aware API Keep-Alive}
\label{sec:keepalive}

LLM API providers retain prompt prefixes for a fixed time-to-live (TTL)~\cite{claude_prompt_caching_retention_5min,openai_prompt_caching}.
Requests arriving before the moment when TTL expires reuse the cached prefix at a discounted rate; requests arriving afterward pay the full input-token price.
We demonstrate a harness-side \textit{keep-alive} policy (\Cref{fig:use-cases}) that uses the idle-time predictor and last-turn predictor (\Cref{sec:idle-predictor,app:last-turn-prediction-setup}) to preserve prefixes across turns.
At each turn boundary, the idle duration is predicted; if the gap is expected to exceed the TTL and the session is unlikely to end, the harness sends a dummy refresh request to refresh the TTL counter.

We evaluate a 300s TTL, matching Claude's default five-minute retention~\cite{claude_prompt_caching_retention_5min}, with refreshes every 290s.
Refreshing stops when the predicted chance that the user returns, scaled by the saving a later hit can still recover, falls below a threshold.
We assume a cached request costs 0.10 of an uncached one, matching major providers' cache-read pricing; retention and pricing details are in \Cref{app:keepalive}.
We compare (1) \uline{\emph{No Keep-Alive}}, which lets the cache expire; (2) \uline{\emph{Predicted Keep-Alive}}, which refreshes based on the prediction; and (3) \uline{\emph{Oracle Keep-Alive}}, which knows the actual gap duration at each boundary and whether it is the last turn.

\myparagraph{Refresh Policy}
The policy considers a turn boundary when the last-turn classifier gives $p_{\mathrm{last}}<\theta_{\mathrm{last}}$.
At the $k$-th refresh time $t=290k$\,s, it issues a refresh only if
\[
\frac{s_k}{s_0}\cdot
\frac{(1-p_{\mathrm{last}})S(t)}
{(1-p_{\mathrm{last}})S(t)+p_{\mathrm{last}}}
\ge \theta_{\mathrm{save}},
\]
where $S(t)$ is the boundary's idle-time survival function (\Cref{subsec:predictor}), $\theta_{\mathrm{save}}$ is the expected-saving threshold, and $s_k=1-0.1\cdot(k+1)$ discounts the remaining savings after successive refreshes.
Refreshes are bounded at nine by the cached-token discount, covering gaps of up to 2,910\,s.
We select $\theta_{\mathrm{last}}=0.4$ and $\theta_{\mathrm{save}}=0.4$ by the highest saving on the training week (June 1--7) and report results on the held-out week (June 8--14), over 1,042,521 turn boundaries including each session's last turn.

\begin{figure}[t]
\centering
\begin{subfigure}[b]{0.49\linewidth}
    \centering
    \includegraphics[width=\linewidth]{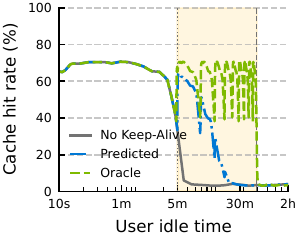}
    \caption{Cache hit rate.}
    \label{fig:keepalive-cache-rate}
\end{subfigure}
\hfill
\begin{subfigure}[b]{0.49\linewidth}
    \centering
    \includegraphics[width=\linewidth]{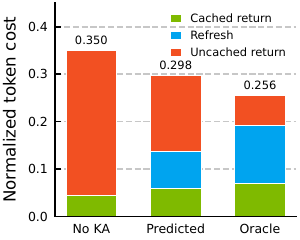}
    \caption{Cost breakdown.}
    \label{fig:keepalive-cost}
\end{subfigure}
\caption{Impacts of the keep-alive policy on (a) cache hit rate and (b) token cost normalized to the uncached baseline.}
\vspace{-10pt}
\label{fig:keepalive}
\end{figure}

\myparagraph{Cache Hit Rate and Cost Savings}
Without refreshes, the cache hit rate falls from approximately 70\% for gaps under one minute to just 4\% for gaps exceeding the 300s TTL (\Cref{fig:keepalive-cache-rate}).
Predicted Keep-Alive maintains a 55\% hit rate for gaps between 5 and 10 minutes, compared with 63\% for the oracle.
Overall, Predicted Keep-Alive reduces normalized cost by 14.9\% at turn boundaries, against the oracle's 26.9\% savings.
Refreshes reduce the per-turn token cost from 0.350 to 0.298 (\Cref{fig:keepalive-cost}).
Classifier errors account for most of the gap to the oracle; \Cref{app:keepalive} analyzes these errors and sweeps both thresholds.

\begin{table}[t]
\centering
\footnotesize
\caption{Predicted Keep-Alive outcomes over all turn boundaries. Each boundary is weighted by its prompt tokens: \emph{Boundaries} is the token-weighted share, and \emph{Avg.\ cost} is the token-weighted mean, normalized to an uncached request. The shares times the costs sum to the policy's overall cost of 0.298.}
\label{tab:keepalive-outcomes}
\begin{tabular}{@{}llrr@{}}
\toprule
\textbf{Actual gap} & \textbf{Refresh} & \textbf{Boundaries} & \textbf{Avg.\ cost} \\
\midrule
$>300$s & Yes & 29.1\% & 0.72 \\
$>300$s & No & 1.4\% & 1.00 \\
$\leq300$s & Yes & 41.9\% & 0.10 \\
$\leq300$s & No & 3.3\% & 0.10 \\
No return & Yes & 15.4\% & 0.18 \\
No return & No & 8.9\% & 0.00 \\
\bottomrule
\end{tabular}
\end{table}

\begin{table}[t]
\centering
\scriptsize
\caption{Keep-alive policy evaluation with a fixed number of refreshes $k$ at every turn boundary, on June 8--14. \emph{Refreshes} is the mean number of refreshes per boundary; \emph{Saving} is the cost reduction over the baseline No Keep-Alive (\%).}
\label{tab:keepalive-baselines}
\setlength{\tabcolsep}{2pt}
\begin{tabular}{@{}lrrrrrrrrr@{}}
\toprule
\textbf{Fixed $k$} & 1 & 2 & 3 & 4 & 5 & 6 & 7 & 8 & 9 \\
\midrule
Refreshes & 0.56 & 1.03 & 1.45 & 1.86 & 2.24 & 2.61 & 2.98 & 3.33 & 3.69 \\
Saving & 11.0 & 10.5 & 6.2 & 0.4 & $-$6.3 & $-$13.4 & $-$20.7 & $-$28.1 & $-$35.6 \\
\bottomrule
\end{tabular}
\end{table}

\myparagraph{Where Savings Come From}
\Cref{tab:keepalive-outcomes} breaks down the token-weighted outcomes.
Correct refreshes on gaps longer than the TTL (29.1\% of boundaries) cut their cost from 1.00 to 0.72, while refreshes on boundaries that never return (15.4\%) add a cost of 0.18.
Only 1.4\% of boundaries are long gaps that the policy misses.

\myparagraph{Comparison with Fixed Refreshes}
A predictor-free policy that refreshes a fixed $k$ times at every boundary saves at most 11.0\% ($k=1$), and becomes net-negative beyond $k=4$ because refreshes are wasted on sessions that never return and on gaps that outlast them (\Cref{tab:keepalive-baselines}).
Predicted Keep-Alive's 14.9\% saving comes from skipping refreshes on likely last turns and stopping early on gaps that are unlikely to return in time.

\section{Idleness-Aware KV Offloading}
\label{sec:offloading}

An idle session's KV cache can remain in GPU HBM, be offloaded to DRAM or SSD, or be evicted.
In this case study, we design an idle-time-aware KV cache residency policy (\Cref{fig:use-cases}, right) that improves eviction decisions (\Cref{sec:offload-design}), then evaluate the policy on two inference engines (\Cref{sec:offload-e2e}) and in a trace-driven simulator (\Cref{sec:offload-ablation}).

\subsection{Idleness-Aware Eviction Policy}
\label{sec:offload-design}

\myparagraph{Motivation}
LRU assumes recently used entries are more likely to be reused soon, an effective heuristic for independent requests.
Agent sessions, however, are largely append-only: each LLM call extends the existing prompt, so resuming a session requires its cached prefix.
Whether that prefix is reused soon depends more on the session's idle duration than on recency alone (\Cref{sec:kv-cache}).
Therefore, predicted idle time can guide which KV prefixes to retain or evict.

\myparagraph{Eviction Policies}
At time \(t\), let \(\mathcal{C}(t)\) be the set of idle prefixes in the cache. For each prefix \(i\), its age \(A_i(t)\) is the time since its last use, and its residual \(R_i(t)\) is the time until its next use.
We compare three policies:
(1) \textbf{LRU:} evict $\arg\max_{i\in\mathcal{C}(t)} A_i(t)$;
(2) \textbf{LRI:} evict $\arg\max_{i\in\mathcal{C}(t)} \widehat{R}_i(t)$;
(3) \textbf{Oracle:} evict $\arg\max_{i\in\mathcal{C}(t)} R_i(t)$.
Here, LRI (\emph{Longest Remaining Idle}) ranks prefixes by predicted time until reuse, \(\widehat{R}_i(t)\), using the idle-time predictor from \Cref{sec:idle-predictor}. LRU uses only past access time, while Oracle has knowledge of future reuse and serves as an ideal reference~\cite{belady1966study}.

\myparagraphnodot{When LRU Fails?}
We measure eviction quality by pairwise agreement with Oracle: the probability that a policy evicts the prefix that will be reused later (defined formally in \Cref{app:recency-metric}).
LRU achieves 0.68--0.73 accuracy across retention horizons, well above random choice but still misranks roughly 27--32\% of candidate pairs.
The gap is because LRU ranks prefixes by \textit{age}, whereas an Oracle ranks them by time until next use, which would yield perfect accuracy.
In production, the two are weakly correlated (Spearman \(\rho=0.40\)--\(0.43\)): older prefixes are somewhat more likely to stay idle longer, but age alone often misranks sessions that are about to resume.
Ranking idle prefixes with the calibrated last-turn probabilities (\Cref{subsec:predictor}) instead raises the pairwise accuracy at a one-hour horizon to 0.78.
This motivates LRI, which incorporates predicted idle time to improve eviction decisions.

\myparagraph{Age-Conditioned Survival Score for LRI}
Ranking prefixes by a point estimate $\widehat{R}_i$ ignores both the long tail of idle durations and the elapsed idle time available at eviction.
We instead rank prefixes by the probability of remaining idle for another fixed horizon $h$, conditioned on the elapsed idle time.
Let $T_i$ denote the idle duration after session $i$'s latest turn.
LRI evicts the prefix with the highest
\begin{equation}
s_i(t)=\Pr[T_i>A_i(t)+h \mid T_i>A_i(t)].
\label{eq:lri-sim}
\end{equation}
Conditioning on elapsed idleness allows LRI to update the ranking as prefixes age, prioritizing sessions likely to remain idle over the next $h$ seconds.

\begin{table*}[t!]
\centering
\scriptsize
\caption{Week-long replay at 200 sessions/h (June 8--14, 33K sessions). We train the model to exclude testing session from training. Mean TTFT in seconds for LRU, LRI, and Oracle (O); LRI's mean-TTFT reduction over LRU (\%) over the week and on each day; DRAM evictions in millions of blocks.}
\label{tab:sim-week}
\setlength{\tabcolsep}{3pt}
\begin{tabular}{@{}l rrr r rrrrrrr rr@{}}
\toprule
& \multicolumn{3}{c}{\textbf{Mean TTFT (s)}} & \multicolumn{8}{c}{\textbf{LRI reduction (\%)}} & \multicolumn{2}{c}{\textbf{Ev.\ (M)}} \\
\cmidrule(lr){2-4}\cmidrule(lr){5-12}\cmidrule(l){13-14}
\textbf{Cap.} & LRU & LRI & O & Week & 6/8 & 6/9 & 6/10 & 6/11 & 6/12 & 6/13 & 6/14 & LRU & LRI \\
\midrule
60K & 6.71 & 5.09 & 3.67 & 24.1 & 21.3 & 22.2 & 29.9 & 22.1 & 31.5 & 17.6 & 14.0 & 238 & 127 \\
90K & 5.32 & 4.16 & 3.23 & 21.8 & 15.4 & 15.7 & 28.1 & 33.3 & 23.3 & 7.6 & 14.7 & 216 & 113 \\
120K & 4.60 & 3.77 & 3.07 & 18.1 & 14.6 & 11.6 & 25.0 & 28.7 & 19.0 & 3.2 & 14.8 & 203 & 105 \\
\bottomrule
\end{tabular}
\end{table*}

\subsection{End-to-end Evaluation}
\label{sec:offload-e2e}

\myparagraph{Setup}
We implement LRU, LRI, and Oracle in vLLM's host-memory KV tier~\cite{kwon2023vllm} and evaluate them by replaying production \platform traces on two 8$\times$A100 nodes~\cite{microsoft_ndma100v4}.
We consider both Prefill-Decode (PD) colocation (Qwen3-32B on 4 A100s) and PD disaggregation (Qwen3-235B-A22B with 8 A100s each for prefill and decode).
We measure prefix-cache hit rate, time to first token (TTFT), and wasted evictions at user-turn boundaries.
Because the traces contain no prompt text, the replay rebuilds each session from its token counts: calls within a session share prefixes as in production, while different sessions share none, and consecutive turns are append-only except at context compaction.
These assumptions follow evidence that production KV hits across users are rare~\cite{wang2025kvcachewild} and that agent harnesses preserve prefixes~\cite{claudecode,openaicodex}; full configurations and justification are in \Cref{app:e2e-details}.

\myparagraph{Policies}
LRU is from vLLM default implementation.
LRI uses the score in \Cref{eq:lri-sim} with a one-hour horizon and predictors in \Cref{subsec:predictor}; replayed sessions are held out (\ie{}, unseen traces during predictor training).
Oracle ranks cached blocks by their true next-use times and provides an upper bound on ranking quality (\Cref{app:e2e-details}).

\begin{figure}[t!]
\centering
\includegraphics[width=\linewidth]{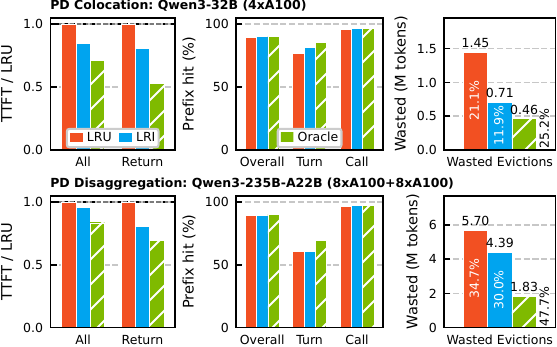}
\caption{End-to-end results comparing LRU, LRI, and Oracle in PD colocation vs. disaggregation setups.
Wasted evictions (in \%) are the fraction of evicted host-tier blocks requested again.
\vspace{-10pt}
}
\label{fig:e2e}
\end{figure}

\myparagraph{Results}
LRI consistently reduces TTFT and wasted evictions across both configurations while maintaining comparable or slightly higher prefix-cache hit rates.
In PD colocation, LRI reduces return TTFT by 20\% and wasted evictions by 51\% (1.45M to 0.71M tokens), lowering the fraction of evicted blocks requested again from 21.1\% to 11.9\%.
In PD disaggregation, it reduces return TTFT by 19\% and wasted evictions by 23\%, with the corresponding fraction decreasing from 34.7\% to 30.0\%.

\begin{figure}[t!]
\centering
\begin{subfigure}[b]{0.49\linewidth}
    \centering
    \includegraphics[width=\linewidth]{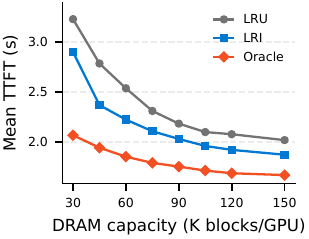}
    \caption{Mean TTFT, DRAM only.}
    \label{fig:sim-ttft-dram}
\end{subfigure}
\hfill
\begin{subfigure}[b]{0.49\linewidth}
    \centering
    \includegraphics[width=\linewidth]{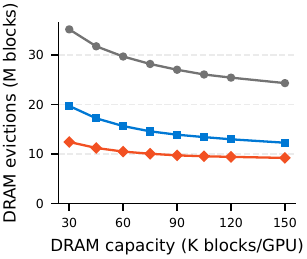}
    \caption{DRAM evictions.}
    \label{fig:sim-evictions}
\end{subfigure}
\\[2pt]
\begin{subfigure}[b]{0.49\linewidth}
    \centering
    \includegraphics[width=\linewidth]{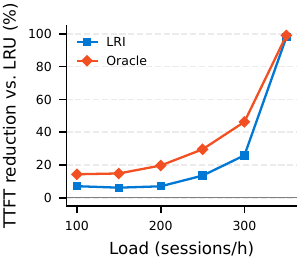}
    \caption{Gain over LRU by load.}
    \label{fig:sim-load-gain}
\end{subfigure}
\hfill
\begin{subfigure}[b]{0.49\linewidth}
    \centering
    \includegraphics[width=\linewidth]{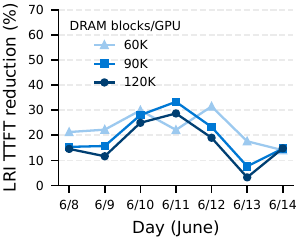}
    \caption{Week-long gain by day.}
    \label{fig:sim-week-gain}
\end{subfigure}
\caption{Ablation study results on KV-cache offloading.}
\vspace{-10pt}
\label{fig:tier-policy}
\end{figure}

\subsection{Ablation and Sensitivity Study}
\label{sec:offload-ablation}

Real deployments span diverse workloads and storage configurations, making it difficult to evaluate eviction policies across this space on physical hardware.
Moreover, user pauses can last from \textit{hours} to \textit{days}, making long-horizon replay expensive.
We therefore build a turn-level simulator that models KV-cache management and request execution, enabling broader capacity and load sweeps, alternative storage configurations, and week-long replays.

\myparagraph{Setup}
We extend Vidur~\cite{agrawal2024vidur} with Cascade~\cite{adnan2026cascade} to simulate LLM serving with hierarchical KV-cache offloading.
We simulate Qwen3-235B-A22B on eight H200 GPUs, using hardware-grounded profiles of prefill/decode operator latency and KV-cache transfer latency.
By default, we replay 4.7K sessions at 200 sessions/h, with 90K DRAM blocks per GPU and no SSD. 
Full simulator details, varied configurations, workload details are in \Cref{app:sim-setup}.

\myparagraph{Impact of DRAM Capacity}
LRI reduces mean TTFT relative to LRU at every tested DRAM capacity from 30K to 150K blocks per GPU (\Cref{fig:sim-ttft-dram}).
The reduction is largest with limited DRAM: 10--15\% at 30K--60K blocks, compared with 7--9\% at 75K--150K blocks.
These gains capture 28--49\% of Oracle's improvement, and LRI reduces P99 TTFT by 0--20\%.
LRI raises the cache hit rate while also evicting 44--50\% fewer DRAM blocks than LRU, versus 62--65\% fewer for Oracle (\Cref{fig:sim-evictions}).

\myparagraph{Impact of Load}
At 90K DRAM blocks per GPU, LRI's benefit increases with load (\Cref{fig:sim-load-gain}), reducing mean TTFT by 6.2--7.1\% at 100--200 sessions/h and 13.5--25.9\% at 250--300 sessions/h.
Under heavier overload scenarios ($>$350 sessions/h), LRI further reduces TTFT by more than 80\% compared with LRU, as queueing increasingly amplifies LRU's eviction degradation.

\myparagraph{Long-Horizon Serving}
We next replay a week of production traces from June 8--14, covering 33K sessions over 168 simulated hours at 200 sessions/h.
LRI reduces mean TTFT by 18--24\% across DRAM capacities of 60K--120K blocks per GPU, outperforms LRU on every day by 3--33\%, and evicts 46--48\% fewer DRAM blocks (\Cref{fig:sim-week-gain,tab:sim-week}).
The largest daily gain occurs on June 11, when LRU's mean TTFT reaches 7.5\,s with 90K blocks.

\begin{figure}[t!]
\centering
\includegraphics[width=0.6\linewidth]{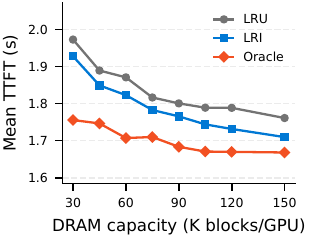}
\caption{Mean TTFT comparison at 200 sessions/h across DRAM capacities with an SSD tier.}
\label{fig:sim-ttft-ssd}
\end{figure}

\myparagraph{Impact of an SSD Tier}
We add an SSD tier (supported in Cascade) that retains every evicted prefix, allowing it to be restored rather than recomputed (\Cref{fig:sim-ttft-ssd}).
All policies then achieve a 70\% hit rate.
At 100--300 sessions/h, LRI improves mean TTFT by only 1.7--4.5\%, primarily by keeping more hits in DRAM.
At 400 sessions/h, SSD prevents LRU's severe overload: its mean TTFT remains 10--16\,s, and LRI reduces it by a further 15--27\%.
With ample SSD capacity, the benefit of better DRAM eviction therefore shrinks except under heavy load.

\myparagraph{Sensitivity to the LRI Horizon}
LRI outperforms LRU at every tested horizon $h$ from five minutes to infinity, with mean-TTFT reductions of 0.6--95.5\%; shorter horizons perform slightly better overall (\Cref{tab:sim-horizon}).
Results for all 72 simulated configurations are in \Cref{app:sim-matrix}.

\section{Related Work}
\label{sec:related}

\myparagraph{Agent Characterization}
AgentSysBench~\cite{chang2026llm} characterize agentic benchmarks such as terminal use, computer use, and auto research.
Agent Arena~\cite{arena2026agentarena} analyzes in-the-wild agent interactions for causal evaluation.
TraceLab~\cite{tracelab} releases 4.3K Claude Code and Codex traces from 43 developers.
\textit{Our study complements these efforts with production-scale \platform traces spanning 819M conversations and 7.4M user, covering diverse coding agent usage and user population.}

\myparagraph{Agent Serving}
Beyond LLM serving, recent work has focused on exploiting agentic workflow structure~\cite{yu2026pythia,luo2025autellix,lin2024parrot,biswas2026sutradhara,guo2026saga,chaudhry2025murakkab,ro2025sherlock,zhang2025jitservesloawarellmserving}.
For example, Pythia~\cite{yu2026pythia} and Autellix~\cite{luo2025autellix} leverage workflow-aware scheduling for agent execution graphs.
Sutradhara~\cite{biswas2026sutradhara} co-designs agent orchestrators and inference engines for tool-augmented workloads.
\textit{Our work on informing KV cache offloading or keep-alive policy with predicted idleness is orthogonal to these improvements.}

\myparagraph{KV Cache Management for Agents}
CacheTTL~\cite{li2025continuum} proposes time-to-live retention of KV cache across tool-call gaps.
CacheWise~\cite{tiwari2026cachewise} improves KV cache reuse with tool-duration prediction.
ICaRus~\cite{woo2026icarus} enables cross-model KV cache sharing.
Our work complements these systems by providing the first \emph{production-scale characterization} of cache lifecycle events, revealing cache reuse opportunities and informing idle-time-aware retention and eviction policies.

\section{Conclusion}
\label{sec:conclusion}

We characterized production coding-agent workloads on \platform, uncovering key properties on agentic workflows, KV-cache lifecycle, and resource idleness.
Based on these findings, we developed a practical idle-time predictor demonstrated in two use cases: harness-side API keep-alive and serving-side KV-cache residency management.
Evaluation shows that keep-alive at turn boundaries reduces token costs by {14.9\%}, while idle-time-aware KV-cache management reduces TTFT by {3--33\%} compared with LRU. 
We open-sourced to the community the traces from our study\footnote{\url{https://github.com/Azure/AzurePublicDataset/blob/master/GitHubCopilotCodingAgentDataset2026.md}} to support future research on efficient agent serving.

\bibliographystyle{mlsys2025}
\bibliography{references}

\clearpage
\appendix
\section{Idle Time Predictor Details}
\label{app:idle-predictor}

\subsection{User-Idleness Prediction Details}
\label{app:user-idle-prediction}

\myparagraph{Features and Model Selection}
We use three feature groups: turn-level metadata (duration, LLM calls, tokens, tool time, failures, and success rate), session-level statistics (turn index, elapsed time, previous idle gap, and average idle duration), and temporal features (e.g., weekday). Session-level features dominate: average idle duration and turn index account for over half of feature importance, followed by the previous idle duration and LLM success rate (\Cref{fig:idle-predictor-feature}).

As shown in \Cref{tab:idle-models}, we compare nine architectures using the same features, predicting whether an idle gap exceeds 60 seconds. All nonlinear models fall within 0.009 ROC-AUC difference from binary LightGBM, whereas logistic regression trails by 0.081.
Adding tool-family indicators, workflow n-grams, session-level workflow statistics, or raw features from the previous one to three turns improves ROC-AUC by at most 0.0031.
We therefore use LightGBM for both quantile and fixed-horizon prediction.

\begin{table}[t]
\centering
\small
\caption{Model selection for predicting idle gaps longer than 60 seconds. $\Delta$ is relative to binary LightGBM.}
\label{tab:idle-models}
\begin{tabular}{@{}lrr@{}}
\toprule
\textbf{Architecture} & \textbf{ROC-AUC} & $\Delta$ \\
\midrule
Binary LightGBM & 0.725 & --- \\
\textbf{Multi-quantile LightGBM} & \textbf{0.724} & \textbf{$-0.001$} \\
RF+XGBoost+LightGBM stacking & 0.725 & +0.000 \\
Three-model ensemble & 0.725 & +0.000 \\
Deeper trees (depth 10) & 0.723 & $-0.002$ \\
AUC-optimized ranker & 0.723 & $-0.002$ \\
Huber regression & 0.716 & $-0.009$ \\
Five-bucket multiclass & 0.716 & $-0.008$ \\
Logistic regression & 0.644 & $-0.081$ \\
\bottomrule
\end{tabular}
\end{table}

For quantile prediction, we train 12 LightGBM regressors on log idle duration, covering quantiles from 0.05 to 0.99, each with 400 trees. For fixed-horizon prediction, a separate binary classifier predicts $T>h$ at turn completion for each horizon $h$ from 30 seconds to 30 minutes. A probability threshold of 0.4 maximizes mean F1 across horizons on June 1--7. Feature importance is measured by the F1 decrease after permuting each feature on the evaluation set (\Cref{fig:idle-predictor-feature}).

\subsection{Tool-Duration Prediction Details}
\label{app:tool-idle-prediction}

\myparagraph{Setup}
We collect training data on the uniformly sampled tool calls (8.34M) from June 1--6 and testing data on June 7 (353K).
Each tool invocation is predicted at dispatch from tool identity, batch shape, turn-start context, and completed prior batches.
The point prediction model is a 500-tree LightGBM regressor~\cite{ke2017lightgbm} with absolute-error loss on $\log(1+\mathrm{duration})$, depth 7, 63 leaves, and learning rate 0.05; separate quantile models target P50--P95, with empirical coverage of 50.8/75.7/90.2/95.2\%.
The baseline predicts each tool's median duration (from the training dataset).
A prediction $\hat d$ is within 2$\times$ of $d$ when $\max(d,\hat d)/\max(\min(d,\hat d),0.1)\leq 2$, in seconds.

\begin{table}[t!]
\centering
\caption{Tool-duration prediction ablation study. Med.\ APE is median absolute percentage error; $\leq$2$\times$ is the within-2$\times$ rate.
}
\label{tab:tool-ablation}
\resizebox{0.95\linewidth}{!}{%
\begin{tabular}{@{}lrrrr@{}}
\toprule
\textbf{Inputs} & \textbf{\#Feat.} & \textbf{Med.\ APE} & \textbf{$\leq$2$\times$} & $R^2$ \\
\midrule
All features & 29 & 54.0\% & 68.9\% & 0.034 \\
Tool name only & 1 & 68.4\% & 61.8\% & -0.002 \\
All but tool name & 28 & 57.7\% & 65.5\% & 0.038 \\
All + final status (oracle) & 30 & 52.9\% & 70.1\% & 0.046 \\
\bottomrule
\end{tabular}
}
\end{table}

\myparagraph{Tool-Prediction Ablation Study}
As shown in \Cref{tab:tool-ablation}, the tool name alone reaches 61.8\% within 2$\times$, 7.1 points below the predictor accuracy with all features.
Removing only the tool name leads to 3.4 points of the accuracy drop, since the tool family and history partly stand in for it.
Knowing the call's final status, which is unavailable at dispatch, adds only 1.2 points.
The longest 1\% of calls contribute 72.4\% of absolute error, which explains the low $R^2$.

\begin{table}[t!]
\centering
\caption{Tool-duration prediction on external traces.}
\label{tab:external-tool-prediction}
\resizebox{0.95\linewidth}{!}{%
\begin{tabular}{@{}llrrr@{}}
\toprule
\textbf{Dataset} & \textbf{Model} & \textbf{MAE (s)} & $R^2$ & \textbf{Within 2$\times$} \\
\midrule
TraceLab & LightGBM & 8.910 & 0.028 & 65.3\% \\
CacheWise & LightGBM & 23.879 & -0.018 & 26.3\% \\
CacheWise & Per-tool median & 20.747 & -0.014 & 41.4\% \\
\bottomrule
\end{tabular}}
\end{table}

\myparagraph{Validation on External Datasets}
We evaluate within-dataset predictors on TraceLab~\cite{tracelab} and the public CacheWise corpus~\cite{tiwari2026cachewise}.
The external predictors use the same objective with 600 trees, depth 8, and 127 leaves, on session-disjoint splits of TraceLab (273K/53K calls) and CacheWise (8.6K/1.3K calls).
Prediction accuracy results are summarized in~\Cref{tab:external-tool-prediction}.
Calls lasting at least 60\,s represent 3.4\% and 7.0\% of test calls, respectively, but contribute 74.3\% and 77.4\% of total tool time and 79.0\% and 79.1\% of absolute prediction error. On CacheWise, the per-tool median outperforms LightGBM. Adding argument-length metadata to TraceLab improves the within-2$\times$ rate only from 65.8\% to 66.0\%.

\subsection{Last-Turn Prediction Setup}
\label{app:last-turn-prediction-setup}
LRI policy uses a last-turn prediction model in addition to the idle duration predictor (\Cref{subsec:predictor}).
A turn is labeled last when no later turn of its session follows in the release.
Both models use the same features (\Cref{subsec:predictor}) as inputs.
The last-turn classifier is a 500-tree binary LightGBM model with depth 6 and 31 leaves, trained with natural class weights so that its outputs are calibrated probabilities.
Its mean predicted probability, 0.252, equals the observed last-turn rate.

\section{Keep-Alive Evaluation Details}
\label{app:keepalive}

\myparagraph{Retention and Pricing Assumptions}
Hosted APIs offer different cache-retention windows. Claude provides five-minute default retention and one-hour retention at a higher cache-write price~\cite{claude_prompt_caching_retention_5min}; OpenAI supports prompt-cache retention of up to 24 hours~\cite{openai_prompt_caching}.
We evaluate the five-minute setting because most requests in the traces use the default retention (\Cref{fig:turn-idle-cache}).
Our keep-alive policy follows Anthropic's five-minute TTL.
Retention differs across providers: OpenAI guarantees 30 minutes~\cite{openai_prompt_caching}, and Amazon Bedrock and Alibaba Cloud's Qwen API guarantee 5 minutes~\cite{bedrock_prompt_caching,alibaba_qwen_context_cache}, each extended by cache hits. The policy still applies as long as a request that reuses a cached prefix extends its retention.
Longer retention (e.g., one hour for Claude) charges a higher write price up front on every turn, whereas keep-alive pays only while the session stays idle; we do not evaluate it here.
Our policy generalizes to other TTLs by adjusting the refresh interval and cost model.
We assume a cached request costs 0.10 relative to an uncached request, matching major LLM API providers' cache-read-to-base-input-price ratio.

\begin{figure}[t]
\centering
\begin{subfigure}[b]{0.49\linewidth}
    \centering
    \includegraphics[width=\linewidth]{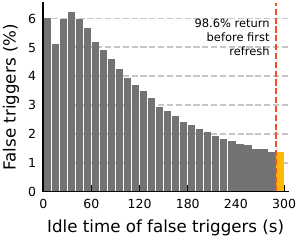}
    \caption{False-trigger returns.}
    \label{fig:keepalive-false-trigger}
\end{subfigure}
\hfill
\begin{subfigure}[b]{0.49\linewidth}
    \centering
    \includegraphics[width=\linewidth]{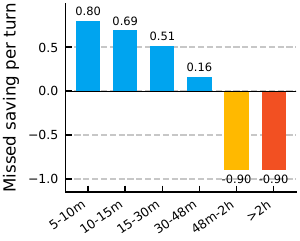}
    \caption{Saving lost to misses.}
    \label{fig:keepalive-missed}
\end{subfigure}
\caption{Classifier errors. (a)~Idle time of turns boundaries with a planned refresh; the dashed line marks the first refresh at 290s. (b)~Average saving lost per missed long gap under oracle, by gap length; beyond 48 minutes, refreshing costs more than it saves.}
\label{fig:keepalive-errors}
\end{figure}

\myparagraph{Cost Accounting}
Keep-alive changes only the prefill (input-token) cost, so we count prefill cost alone.
For LLM call $i$ with $P_i$ prompt tokens, $C_i$ cached tokens, and $U_i=P_i-C_i$ uncached tokens, the estimated prefill cost is
\begin{equation}
\widehat{\mathcal C}_i=
\pi_{\mathrm{uncached},i}U_i+
\pi_{\mathrm{cached},i}C_i.
\label{eq:call-cost}
\end{equation}

Here we mesaure the cost at each turn boundaries with user idleness. Each return and refresh is priced this way and $\pi_{\mathrm{cached}}=0.1\,\pi_{\mathrm{uncached}}$. Costs are normalized by the prefill cost of one uncached request at every turn boundary.

\myparagraph{Detailed Results}
\Cref{fig:keepalive-false-trigger,fig:keepalive-missed} detail where the classifier errors come from, and \Cref{tab:keepalive-sweep} sweeps the two thresholds of Predicted Keep-Alive on the training and evaluation sets.

\begin{table}[t]
\centering
\scriptsize
\caption{Predicted Keep-Alive saving over No Keep-Alive (\%) by last-turn threshold $\theta_{\mathrm{last}}$ and expected-saving threshold $\theta_{\mathrm{save}}$. Top: training set (June 1--7), used for selection. Bottom: evaluation set (June 8--14). Bold is the selected configuration; $\theta_{\mathrm{last}}=0.6$ matches no last-turn threshold.}
\label{tab:keepalive-sweep}
\setlength{\tabcolsep}{3.5pt}
\begin{tabular}{@{}lrrrrrrrr@{}}
\toprule
$\theta_{\mathrm{last}}$ & \multicolumn{8}{c}{$\theta_{\mathrm{save}}$} \\
& 0.0 & 0.1 & 0.2 & 0.3 & 0.4 & 0.5 & 0.6 & 0.7 \\
\midrule
\multicolumn{9}{@{}l}{\emph{Training set (June 1--7)}} \\
0.1 & 3.0 & 4.5 & 5.3 & 5.8 & 5.9 & 5.6 & 4.8 & 3.1 \\
0.2 & 1.0 & 7.8 & 11.3 & 13.1 & 13.7 & 12.6 & 9.7 & 3.9 \\
0.3 & $-$6.6 & 5.8 & 11.9 & 15.2 & 16.1 & 14.4 & 10.1 & 3.9 \\
0.4 & $-$11.3 & 4.3 & 11.7 & 15.5 & \textbf{16.4} & 14.5 & 10.1 & 3.9 \\
0.5 & $-$12.5 & 4.0 & 11.6 & 15.5 & 16.4 & 14.5 & 10.1 & 3.9 \\
None & $-$28.7 & 3.9 & 11.6 & 15.5 & 16.4 & 14.5 & 10.1 & 3.9 \\
\midrule
\multicolumn{9}{@{}l}{\emph{Evaluation set (June 8--14)}} \\
0.1 & 2.7 & 4.0 & 4.7 & 5.2 & 5.3 & 5.0 & 4.3 & 2.8 \\
0.2 & $-$0.9 & 5.8 & 9.5 & 11.7 & 12.5 & 11.7 & 9.2 & 3.7 \\
0.3 & $-$11.2 & 2.2 & 9.1 & 13.1 & 14.6 & 13.5 & 9.6 & 3.7 \\
0.4 & $-$17.7 & $-$0.2 & 8.4 & 13.2 & \textbf{14.9} & 13.6 & 9.6 & 3.7 \\
0.5 & $-$19.2 & $-$0.7 & 8.3 & 13.2 & 14.9 & 13.6 & 9.6 & 3.7 \\
None & $-$35.6 & $-$0.8 & 8.2 & 13.2 & 14.9 & 13.6 & 9.6 & 3.7 \\
\bottomrule
\end{tabular}
\end{table}

\section{KV Offloading Details}

\subsection{Eviction Metrics and Notation}
\label{app:recency-metric}

For each idle prefix \(i\in\mathcal{C}(t)\), we define its age and residual as:
\begin{equation}
A_i(t)=t-t_i^{\mathrm{last}},\qquad
R_i(t)=t_i^{\mathrm{next}}-t,
\label{eq:age-residual}
\end{equation}
where \(R_i(t)=+\infty\) if the prefix is never reused. Let \(\widehat{R}_i(t)\) denote the predicted residual from the idle-time predictor.

We compare eviction policies by their pairwise agreement with Oracle. A policy \(\pi\) assigns each prefix a score \(s_\pi\) and evicts the prefix with the highest score. For LRU, \(s_{\mathrm{LRU}}=A\); for LRI, \(s_{\mathrm{LRI}}=\widehat{R}\).
We define accuracy as:
\begin{equation}
\mathrm{Acc}(\pi)
=\Pr\bigl(R_i>R_j\mid s_\pi(i)>s_\pi(j),\ R_i\ne R_j\bigr).
\label{eq:lru-auc}
\end{equation}
This measures how often the policy evicts the prefix that will be reused later when two candidates are compared.
Accuracy is 1 for perfect agreement with Oracle and 0.5 for random ranking; values below 0.5 indicate worse-than-random ranking.
We also measure the Spearman correlation \(\rho\) between prefix age \(A\) and residual \(R\) across the candidate population.
A high positive correlation indicates that older prefixes tend to be reused later, favoring LRU; a weak or negative correlation indicates that age is a less reliable proxy for future reuse.
Comparing \(\mathrm{Acc}(\mathrm{LRI})\) with \(\mathrm{Acc}(\mathrm{LRU})\) isolates whether predicted idle time improves eviction ranking over recency alone, which turns out to be the scenarios where \(\rho\) is low (\eg{}, \(\rho=0.40\)--\(0.43\) in production).

\begin{table}[t]
\centering
\small
\caption{Notation for eviction analysis.}
\label{tab:recency-terms}
\begin{tabular}{@{}ll@{}}
\toprule
\textbf{Symbol} & \textbf{Definition} \\
\midrule
\(\mathcal{C}(t)\) & Idle prefixes resident at time \(t\) \\
\(A_i(t)\) & Time since prefix \(i\) was last used \\
\(R_i(t)\) & Time until prefix \(i\) is next used; \(+\infty\) if never \\
\(\widehat{R}_i(t)\) & Predicted time until next use \\
\(\rho\) & Spearman correlation between age and residual \\
\(\mathrm{Acc}(\pi)\) & Pairwise eviction accuracy of policy \(\pi\) \\
\bottomrule
\end{tabular}
\vspace{-15pt}
\end{table}

\subsection{End-to-end Setup Details}
\label{app:e2e-details}

\myparagraph{Setup}
We use vLLM's host-memory KV tier~\cite{kwon2023vllm}, which writes a block to host memory when GPU memory evicts it.
Each run replays complete sessions with a one-hour warm-up and a three-hour measurement window.
Both models natively support 32,768-token contexts; we extend them to 262,144 tokens.
This does not affect our measurements because the replay fixes each LLM call's output length to the trace's, so the generated text does not change the workload.
A 262K-token context covers the longest prompts of the sampled sessions (228K tokens for the single engine and 259K for 1P1D).

\myparagraph{Trace Reconstruction}
The trace records token counts and timestamps but no prompt text, so the replay rebuilds each session from its counts.
For each session and model, it draws one pseudo-random token-ID sequence seeded by the session ID and sends each LLM call the first $n$ tokens of that sequence, where $n$ is the call's recorded prompt length.
Calls in a session therefore share prefixes as in production, while different sessions share none.
Each call generates exactly its recorded number of output tokens.
Within a session, calls run serially in timestamp order; after each call completes, the replay waits for the recorded tool idle time before the next call and for the recorded user pause duration before the next turn, so measured GPU latency replaces the recorded LLM duration from the traces.

\myparagraph{Prefix Cache Modeling}
(1) We do not model prefix sharing across sessions: the trace records no prompt content that would identify shared prefixes, coding-agent sessions are long and come from many different users, and production KV-cache hits across users are rare~\cite{wang2025kvcachewild}. We believe this assumption is reasonable because some recent inference deployments do not specifically optimize for shared system prompts.
By default, Azure OpenAI caches only the prefix ending at a request's latest message, so sessions do not reuse a shared system prompt without an explicit breakpoint~\cite{azure_openai_prompt_caching}.
CachedAttention likewise manages KV cache per conversation session, using all of a session's KV cache as the minimal eviction and fetching unit~\cite{gao2024cost}.
(2) We also model consecutive turns as append-only.
We verify this by comparing token counts between consecutive turns across the population.
We also verify that the \platform coding-agent harness does not modify the prompt prefix except during context compaction, which is rare; because compaction rewrites almost the entire history except the system prompt, we reset the prefix cache to zero after each compaction.
We believe this assumption is reasonable because most harnesses~\cite{claudecode, openaicodex} strive to preserve the prefix cache rather than invalidate it. 

\myparagraph{Policy}
LRI uses the last-turn classifier and idle model of \Cref{app:last-turn-prediction-setup}.
At each turn's completion, it records the probability $\ell_i$ that the turn is the session's last and the idle-time distribution $F_i$ otherwise, as in \Cref{eq:lri-sim}.
The eviction heaps are rebuilt from current scores every 10 seconds.
LRU evicts the least recently used blocks. Oracle evicts the blocks that have longest gap to be next reused. 

\paragraph{Single-Engine Setup}
\label{app:e2e-single}

We serve Qwen3-32B in BF16 on four of the eight A100-80GB GPUs with 96 AMD EPYC 7V12 (Rome) cores, using tensor parallelism of four.
After 16.7\,GB of model weights per GPU, the engine reserves 58.4\,GB per GPU for the KV cache (234\,GB in total), and host memory holds 117\,GB.
To make sure every run completed without request errors and with a stable queue, we scale the arrival rate to around 32 session per hour.

\paragraph{1P1D Setup}
\label{app:e2e-pd}

We serve Qwen3-235B-A22B in FP8 on two VMs, each with eight A100-80GB GPUs and 96 AMD EPYC 7V12 (Rome) cores: a prefill instance on one VM and a decode instance on the other, each using tensor parallelism of eight with expert parallelism.
After 30.4\,GB of model weights per GPU, each instance reserves 43.8\,GB per GPU for the KV cache (350\,GB per instance).
The instance also holds a 175\,GB host-memory tier and runs the eviction policy. Two nodes are connected with Infiniband network.
A proxy runs each LLM call on the prefill instance with a one-token output; the decode instance then pulls the prompt KV over NIXL and generates the response, including its first token.
The workload is about 64 sessions per hour, a rate at which every run completed with a stable queue.

\subsection{Simulator}
\label{app:sim-setup}

\myparagraph{Implementation Details}
Our simulator builds on Cascade~\cite{adnan2026cascade}, which extends Vidur's discrete-event model of an LLM engine and its vLLM scheduler~\cite{agrawal2024vidur} with a hierarchical KV-cache spanning across HBM, DRAM, and SSD.
We adopt a C++ reimplementation of Cascade, which runs faster and scales to multi-rack deployments.
Since the trace anonymizes model identities, we configure Qwen3-235B-A22B on eight H200 GPUs (tensor parallelism).
With Vidur, we model prefill and decode latency using operator-level profiles collected by running LLM model operators on real H200 hardware.
This hardware-grounded profiling provides high-fidelity latency estimates under the model configuration.

\myparagraph{KV-Cache Model}
In Cascade's KV-cache model, each GPU holds a prefix cache of 16-token blocks indexed by logical prefix hashes; as in vLLM, a request reuses cached blocks up to the first missing block.
Blocks evicted from HBM are written to DRAM and, optionally, SSD, while requests hitting a lower tier must load the corresponding blocks before prefill.
Each GPU has 13,580 HBM blocks, with DRAM capacity varying across experiments; the optional SSD tier holds 12M blocks.
We model transfer latency as a fixed startup cost plus a bandwidth-dependent term, parameterized by measurements on the actual GPU--CPU and CPU--SSD transfer links.
The resulting models are approximately $10\,\mu\text{s}+\text{bytes}/50\,\text{GB/s}$ for DRAM and $100\,\mu\text{s}+\text{bytes}/12\,\text{GB/s}$ for SSD.
Thus, both computation and data movement are grounded in real-hardware measurements, providing high-fidelity inputs to the simulator.
The transfer models achieve a mean absolute percentage error of $<${2\%} across evaluated block sizes.
LRU, LRI, and Oracle differ only in their DRAM and SSD block eviction policies.

\myparagraph{Workload and Arrival Process}
We replay coding-agent traces using recorded tool-driven idle times within turns and user-driven idle times across turns, while preserving system metadata to predict idle times. We use same prefix cache modeling in end-to-end evaluation. 

\begin{table*}[t]
\centering
\small
\caption{Sensitivity of LRI to the horizon $h$ in \Cref{eq:lri-sim} in terms of mean-TTFT reduction over LRU (\%), by load (sessions per hour) and DRAM blocks per GPU, without and with an SSD tier. $h=\infty$ ranks prefixes by the probability that the session ends.}
\label{tab:sim-horizon}
\begin{tabular*}{\textwidth}{@{\extracolsep{\fill}}lrrrrrr@{}}
\toprule
& \multicolumn{3}{c}{\textbf{250 sessions/h, DRAM}} & \multicolumn{2}{c}{\textbf{90K, DRAM}} & \textbf{SSD} \\
\cmidrule(lr){2-4}\cmidrule(lr){5-6}\cmidrule(l){7-7}
$h$ & 60K & 90K & 120K & 130 sessions/h & 380 sessions/h & 250 sessions/h, 90K \\
\midrule
5\,min & 21.2 & 13.3 & 11.6 & 6.7 & 95.5 & 3.5 \\
15\,min & 22.4 & 13.2 & 12.8 & 6.7 & 95.2 & 3.9 \\
30\,min & 22.0 & 12.3 & 9.9 & 6.7 & 94.4 & 2.7 \\
1\,h & 20.8 & 12.9 & 10.5 & 6.6 & 94.0 & 1.8 \\
2\,h & 19.6 & 11.9 & 9.8 & 6.0 & 91.7 & 2.0 \\
4\,h & 19.4 & 11.2 & 8.3 & 6.3 & 90.7 & 2.1 \\
8\,h & 19.0 & 11.4 & 8.1 & 6.3 & 90.2 & 1.9 \\
$\infty$ & 18.2 & 9.2 & 7.2 & 6.4 & 89.8 & 0.6 \\
\bottomrule
\end{tabular*}
\end{table*}

\myparagraph{Full Simulator Results}
\label{app:sim-matrix}
\Cref{tab:sim-matrix-metadata} reports the 72 configurations using the default prefix-matching rule.

\Cref{tab:sim-horizon} varies the LRI horizon from five minutes to infinity.

\begin{table*}[t]
\centering
\scriptsize
\caption{Full simulation results. \textbf{Load} is sessions per hour; \textbf{Cap.}\ is DRAM blocks per GPU; \textbf{Mean} and \textbf{P99} are TTFT in seconds; \textbf{Hit} is the cache hit rate in percent; \textbf{Ev.}\ is DRAM evictions in millions of blocks. O is Oracle.}
\label{tab:sim-matrix-metadata}
\setlength{\tabcolsep}{3pt}
\begin{tabular}{@{}lll rrr rr rrr rrr@{}}
\toprule
& & & \multicolumn{3}{c}{\textbf{Mean TTFT (s)}} & \multicolumn{2}{c}{\textbf{P99 TTFT (s)}} & \multicolumn{3}{c}{\textbf{Hit (\%)}} & \multicolumn{3}{c}{\textbf{Ev.\ (M)}} \\
\cmidrule(lr){4-6}\cmidrule(lr){7-8}\cmidrule(lr){9-11}\cmidrule(l){12-14}
\textbf{Tier} & \textbf{Load} & \textbf{Cap.} & LRU & LRI & O & LRU & LRI & LRU & LRI & O & LRU & LRI & O \\
\midrule
DRAM & 100 & 60K & 1.38 & 1.29 & 1.16 & 13.4 & 13.0 & 63.4 & 65.7 & 69.3 & 12.9 & 6.5 & 4.3 \\
DRAM & 100 & 90K & 1.31 & 1.21 & 1.12 & 13.1 & 12.6 & 65.2 & 67.7 & 69.9 & 12.0 & 5.8 & 4.1 \\
DRAM & 100 & 120K & 1.27 & 1.18 & 1.12 & 12.9 & 12.5 & 66.2 & 68.5 & 69.9 & 11.4 & 5.5 & 4.0 \\
DRAM & 200 & 60K & 2.54 & 2.22 & 1.85 & 21.5 & 18.1 & 58.6 & 61.9 & 67.9 & 29.7 & 15.6 & 10.5 \\
DRAM & 200 & 90K & 2.18 & 2.03 & 1.75 & 18.0 & 17.9 & 61.9 & 64.9 & 69.3 & 27.0 & 13.9 & 9.7 \\
DRAM & 200 & 120K & 2.08 & 1.92 & 1.69 & 17.4 & 16.7 & 63.8 & 66.5 & 69.7 & 25.4 & 12.9 & 9.4 \\
DRAM & 300 & 60K & 8.11 & 5.38 & 3.12 & 74.6 & 53.9 & 53.0 & 57.3 & 65.9 & 50.4 & 27.8 & 18.1 \\
DRAM & 300 & 90K & 5.30 & 3.93 & 2.85 & 51.8 & 34.1 & 57.9 & 61.7 & 68.0 & 44.8 & 23.9 & 16.4 \\
DRAM & 300 & 120K & 4.18 & 3.44 & 2.72 & 37.9 & 31.0 & 60.7 & 64.1 & 69.0 & 41.4 & 21.9 & 15.6 \\
DRAM & 400 & 60K & 1615 & 782 & 16.23 & 3307 & 1918 & 17.2 & 35.0 & 62.8 & 119.9 & 44.6 & 27.4 \\
DRAM & 400 & 90K & 1502 & 223 & 9.79 & 3305 & 786 & 20.7 & 51.7 & 66.2 & 114.6 & 43.8 & 24.8 \\
DRAM & 400 & 120K & 1158 & 24.53 & 7.34 & 2737 & 170 & 26.7 & 60.7 & 67.8 & 105.6 & 34.5 & 23.2 \\
\midrule
DRAM+SSD & 100 & 60K & 1.19 & 1.17 & 1.12 & 12.1 & 12.1 & 69.9 & 69.9 & 69.8 & 12.9 & 6.5 & 4.3 \\
DRAM+SSD & 100 & 90K & 1.17 & 1.14 & 1.12 & 12.1 & 12.0 & 69.9 & 69.9 & 69.9 & 12.0 & 5.8 & 4.1 \\
DRAM+SSD & 100 & 120K & 1.16 & 1.13 & 1.12 & 12.1 & 12.0 & 69.9 & 69.8 & 69.9 & 11.4 & 5.5 & 4.0 \\
DRAM+SSD & 200 & 60K & 1.87 & 1.82 & 1.71 & 15.3 & 15.3 & 70.0 & 70.0 & 70.0 & 29.7 & 15.6 & 10.4 \\
DRAM+SSD & 200 & 90K & 1.80 & 1.77 & 1.68 & 14.8 & 15.0 & 70.0 & 70.0 & 70.0 & 27.0 & 13.9 & 9.7 \\
DRAM+SSD & 200 & 120K & 1.79 & 1.73 & 1.67 & 14.8 & 14.8 & 70.0 & 70.0 & 70.0 & 25.4 & 12.9 & 9.4 \\
DRAM+SSD & 300 & 60K & 3.11 & 2.97 & 2.65 & 26.0 & 25.6 & 70.2 & 70.1 & 70.1 & 49.9 & 27.6 & 18.1 \\
DRAM+SSD & 300 & 90K & 2.92 & 2.82 & 2.60 & 24.5 & 23.8 & 70.2 & 70.1 & 70.2 & 44.7 & 23.9 & 16.4 \\
DRAM+SSD & 300 & 120K & 2.84 & 2.71 & 2.58 & 23.2 & 22.9 & 70.2 & 70.2 & 70.1 & 41.4 & 21.8 & 15.6 \\
DRAM+SSD & 400 & 60K & 16.39 & 11.97 & 7.60 & 151 & 126 & 70.1 & 70.0 & 70.0 & 77.8 & 44.0 & 27.2 \\
DRAM+SSD & 400 & 90K & 11.48 & 9.51 & 6.75 & 131 & 113 & 70.1 & 70.1 & 70.0 & 67.4 & 37.8 & 24.7 \\
DRAM+SSD & 400 & 120K & 9.71 & 8.21 & 6.31 & 120 & 94.4 & 70.1 & 70.0 & 70.0 & 61.7 & 33.8 & 23.2 \\
\bottomrule
\end{tabular}
\end{table*}

\end{document}